\documentclass[pdflatex,sn-apa]{sn-jnl}%

\usepackage{graphicx}%
\usepackage{multirow}%
\usepackage{amsmath,amssymb,amsfonts}%
\usepackage{amsthm}%
\usepackage{mathrsfs}%
\usepackage[title]{appendix}%
\usepackage{xcolor}%
\usepackage{textcomp}%
\usepackage{manyfoot}%
\usepackage{booktabs}%
\usepackage{algorithm}%
\usepackage{algorithmicx}%
\usepackage{algpseudocode}%
\usepackage{listings}%

\theoremstyle{thmstyleone}%

\theoremstyle{thmstyletwo}%

\theoremstyle{thmstylethree}%

\usepackage{subcaption}
\usepackage{cleveref}

\usepackage{tikz}
\usepackage{pgfplots}
\usepgfplotslibrary{groupplots}
\usetikzlibrary{spy,calc,positioning,patterns,fit,calc,shapes,pgfplots.groupplots,plotmarks,arrows.meta,shapes.geometric,matrix}
\pgfplotsset{compat=newest}
\usepackage{xcolor}

\usepackage{xspace}

\makeatletter
\DeclareRobustCommand\onedot{\futurelet\@let@token\@onedot}
\def\@onedot{\ifx\@let@token.\else.\null\fi\xspace}

\def\ie{\emph{i.e}\onedot}

\makeatother

\usepackage{acronym}

\newacro{cer}[CER]{Character Error Rate}
\newacro{cnn}[CNN]{Convolutional Neural Network}
\newacro{dm}[DM]{Diffusion Model}
\newacro{fid}[FID]{Fréchet Inception Distance}
\newacro{gan}[GAN]{Generative Adverserial Network}
\newacro{htr}[HTR]{Handwritten Text Recognition}
\newacro{hwd}[HWD]{Handwriting Distance}
\newacro{kid}[KID]{Kernel Inception Distance}
\newacro{kl}[KL]{Kullback-Leibler}
\newacro{ldm}[LDM]{Latent Diffusion Model}
\newacro{mae}[MAE]{Mean Absolute Error}
\newacro{mse}[MSE]{Mean Squared Error}
\newacro{vae}[VAE]{Variational Autoencoder}
\newacro{wi}[WI]{Writer Identification}

\begin{document}

\title[Zero-Shot Paragraph-level Handwriting Imitation]{Zero-Shot Paragraph-level Handwriting Imitation with Latent Diffusion Models}

\author*[1]{\fnm{Martin} \sur{Mayr}}\email{martin.mayr@fau.de}

\author[1]{\fnm{Marcel} \sur{Dreier}}\email{marcel.dreier@fau.de}

\author[1]{\fnm{Florian} \sur{Kordon}}\email{florian.kordon@fau.de}

\author[1]{\fnm{Mathias} \sur{Seuret}}\email{mathias.seuret@fau.de}

\author[2]{\fnm{Jochen} \sur{Zöllner}}\email{jochen.zoellner@planet-ai.de}

\author[1]{\fnm{Fei} \sur{Wu}}\email{river.wu@fau.de}

\author[1]{\fnm{Andreas} \sur{Maier}}\email{andreas.maier@fau.de}

\author[1]{\fnm{Vincent} \sur{Christlein}}\email{vincent.christlein@fau.de}

\affil[1]{\orgdiv{Pattern Recognition Lab}, \orgname{Friedrich-Alexander-Universität Erlangen-Nürnberg (FAU)}, \orgaddress{%
\city{Erlangen}, 
\country{Germany}}}

\affil[2]{\orgname{Planet AI GmbH}, 
\orgaddress{%
\city{Rostock}, 
\country{Germany}}}

\abstract{The imitation of cursive handwriting is mainly limited to generating handwritten words or lines. Multiple synthetic outputs must be stitched together to create paragraphs or whole pages, whereby consistency and layout information are lost. To close this gap, we propose a method for imitating handwriting at the paragraph level that also works for unseen writing styles. 
Therefore, we introduce a modified latent diffusion model that enriches the encoder-decoder mechanism with specialized loss functions that explicitly preserve the style and content.
We enhance the attention mechanism of the diffusion model with adaptive 2D positional encoding and the conditioning mechanism to work with two modalities simultaneously: a style image and the target text. This significantly improves the realism of the generated handwriting.
We set a new benchmark in our comprehensive evaluation, achieving 61\,\% mAP and 56\,\% top-1 accuracy in style preservation, significantly outperforming the previous best method (37\,\% mAP, 30\,\% top-1). 
We are making our code publicly available for reproducibility, supporting research in this area and research into potential countermeasures: \url{https://github.com/M4rt1nM4yr/paragraph_handwriting_imitation_ldm}
}


\maketitle

\begin{figure}[h]
    \centering
    \includegraphics[width=\textwidth]{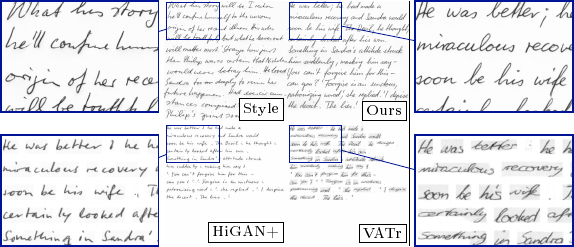}
    \caption{Compared to existing methods, our approach produces more realistic synthetic handwritten paragraphs in a specific style.}
    \label{fig:teaser}
\end{figure}

\section{Introduction}\label{sec:intro}

Advancements in handwritten text generation and imitation hold significant promise for preserving the personal qualities of handwriting, which health conditions or injuries may compromise~\citep{bisio2016kinematics}. These techniques function as a digital preservation mechanism, ensuring continuity of expression for individuals facing physical constraints or other kinds of restrictions. However, as with other deep learning paradigms, their effectiveness depends on the variety and size of suitable training data. Notably, current datasets present challenges, including biases towards certain writing styles, under-representation of languages, and limitations of common data augmentation for tasks such as \ac{htr}, \ac{wi}, and Visual Question Answering (VQA) in document contexts.
Research to date has primarily concentrated on text generation at the word or line level due to the inherent complexities of processing larger coherent textual and visual entities. This focus, however, has led to shortcomings in producing consistent and realistic outputs at the paragraph level -- a prerequisite for practical applicability in many real-world applications, such as personalized text messages or rendering writings in different languages. 

Our study introduces a novel method for paragraph-level handwriting imitation that employs an adapted version of \acfp{ldm}~\citep{rombach2022ldm}. 
Diffusion models typically require large amounts of data and significant computational resources, especially when handling high-resolution inputs such as 768 × 768 images in our case. To address this, the \ac{ldm} uses a \ac{vae} to condense essential information into a more compact latent space, reducing computational demands while preserving crucial details.
We enhance the encoder-decoder framework with style and content preservation loss terms, improving the fidelity and compression of the latent representation. 
Furthermore, we incorporate global positional information and cross-attention mechanisms within the Denoising U-Net architecture in latent space. These enhancements lead to more realistic paragraph generations.
Evaluated as a zero-shot algorithm, our method demonstrates robustness
and generalizability across previously unseen handwriting styles and writers,
significantly outperforming existing methods in synthetic paragraph matching.
The method achieves a top-1 score of over 54\,\% when matching the synthetic
paragraphs with genuine data, almost twice as many percentage points as the
second-best approach.

In summary, our contributions to the field of handwriting generation and imitation include:
(1)~\emph{End-to-end framework} for imitating entire paragraphs of handwritten text. Our method preserves the individual's unique writing style and maintains the original layout, representing a significant step forward in the fidelity of imitated generative handwriting.
(2)~\emph{Refined encoder-decoder stage} by incorporating specialized loss terms that target content and style preservation. We show that these auxiliary losses enhance the generation quality and the latent compression ratio.
(3)~\emph{Improved conditioning} process by integrating the writing style with the target text and employing cross-attention to incorporate this combined information into the Denoising U-Net.
(4)~\emph{Ranked sampling}: Based on the variance within the sampling process, we introduce a ranking scheme that simultaneously considers content and style preservation. 
(5)~\emph{Qualitative and quantitative analyses} show that our method surpasses current state-of-the-art imitation methods by a large margin, considering the combination of image generation, style preservation, and content preservation.

\section{Related Work}
\label{sec:related_work}

In this work, we generate handwritten text solely from images without relying on additional modalities such as the methods that utilize online trajectories~\citep{graves2014generating,mayr2020spatio,aksan2018deepwriting,chang2022controllable,luhman2020diffusion}.  
Unlike online handwriting methods, handwritten text images are widely available, offering broader application possibilities. Various strategies are employed at different levels of detail. Techniques commonly applied to Chinese handwriting or character-specific methods are infrequently used for cursive handwriting in Western scripts~\citep{dai2023disentangling,tang2022fewshot,huang2022agtgan}.
GANwriting~\citep{kang2020ganwriting} generated images on a word level based on a few style samples. They extended their approach to also work on full lines~\citep{kang2021content}. As the name suggests, they use a \ac{gan}. Like most approaches, the style samples and target texts are encoded initially. An upsampling generator produces the output image based on the concatenated style and text information, while AdaIN~\citep{huang2017adain} is used for guidance. 
The two default \ac{gan} losses (discriminator and generator loss) are extended by the domain-specific feedback of the writer and recognition losses. 
Similarly, ScrabbleGAN~\citep{fogel2020scrabble} and TS-GAN~\citep{davis2020text} applied a GAN to generate text lines, but the former one only used \ac{htr} feedback as an extra loss term, and the latter one added a space predictor to space the text for the generator.
SmartPatch~\citep{mattick2021smartpatch} and SLOGAN~\citep{luo2023slogan} added character feedback to improve the results on the stroke level. 
By contrast, HiGAN+~\citep{gan2022higanplus} applies a patch discriminator with a fixed grid of extracted patches but additionally regularizes the style by reconstructing the style vector, which is uniformly sampled, like in JokerGAN~\citep{zdenek2021jokergan}.
With JokerGAN++~\citep{zdenek2023charactergan} they exchanged their style encoder with a ViT~\citep{dosovitskiy2021vit}.
To further increase realism in the outputs, transformer models~\citep{bhunia2021handwriting} as generators and visual archetypes~\citep{pippi2023archetypes} are applied. 

Recent advancements in \acp{dm} in the field of computer vision~\citep{sohl2015dm,song2021denoising,ho2020dm,yang2023dmsurvey} have also influenced the research in handwriting generation. This progress has rendered various
customized loss terms obsolete~\citep{ding2023improving,zhu2023diff,nikolaidou2023wordstylist}. 
Most of these diffusion methods are constrained to pre-existing writing styles since they incorporate the writer ID as a style input in their designs. Hence they cannot generalize to unseen styles. However, CTIG-DM~\citep{zhu2023diff} differentiates between the style of the writer and the style of the image, with the latter primarily focusing on texture and colour.
Interestingly, GC-DDPM~\citep{ding2023improving} also incorporates visual archetypes into their approach for more stability, similar to VATr~\citep{pippi2023archetypes}. 
Moreover, \citep{nauman2024stylus} applied diffusion models to German text data. While all of these diffusion-based methods operate at the word level, our approach directly generates entire paragraphs.
This is achieved via an adapted \ac{ldm}, as described in \cref{sec:methodology}.
Additionally, our ranked resampling method, detailed in \cref{sec:rerank}, accounts for both content and style, whereas Ding et al.'s approach solely ranks based on character correctness.

\section{Methodology}
\label{sec:methodology}

\acp{dm}~\citep{sohl2015dm,song2021denoising,ho2020dm,yang2023dmsurvey} are ubiquitous for image generation, but their application to high-resolution images is data- and resource-intensive. To mitigate this, \acp{ldm} train a diffusion model in a compressed latent space, accessible from the pixel space with an encoder-decoder pair~\citep{ramesh2021dalle,rombach2022ldm}. 
Further, despite impressive results on natural images, \acp{dm} often lack the capabilities to produce realistic-looking text. Only TextDiffuser~\citep{chen2023textdiffuser} produces realistic scene text images, mainly limited to fonts.
Therefore, we applied several modifications as described below to be able to generate handwritten paragraphs.

\begin{figure}[t]
    \centering
    \includegraphics[width=0.95\textwidth]{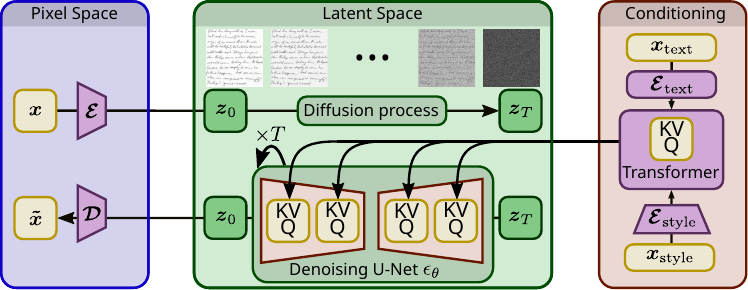}
    \caption{Method overview. We transfer the handwritten paragraphs in and out of latent space via encoder $\mathcal{E}$ and decoder $\mathcal{D}$. The Denoising U-Net $\epsilon_{\Theta}$ is trained in latent space and conditioned with cross-attention. As conditioning information, we have two inputs: (1) a style image $x_{\text{style}}$, which we encode with a shallow CNN $\mathcal{E}_{style}$, and (2) a target text $x_{\text{text}}$, which we embed into feature space. We fuse both modalities with a transformer and forward them as a stylized embedding into the Denoising U-Net via cross-attention.}
    \label{fig:method_overview}
\end{figure}

Given a style image $x_{\text{style}}$ and a target text $x_{\text{text}}$, the task of handwritten text imitation can be described as to produce an output image $\Tilde{x}$ mimicking the style with the given content. 
For training, $x$ and $x_{\text{style}}$ are from the same writer but, if possible, different paragraphs. 
\Cref{fig:method_overview} visualizes the building blocks to solve this task, which are elaborated in the upcoming subsections. 

\subsection{Encoder-Decoder Stage}
\label{sec:methodology:encdec}

First, the translation into the latent representation is applied, \ie, $z = \mathcal{E}(x)$ where $\mathcal{E}$ denotes the encoder. 
The images $\Tilde{x} = \mathcal{D}(x)$ are reconstructed with decoder $\mathcal{D}$.
This step's important property is condensing the image information into a compressed representation. 
This is typically accomplished by reducing the spatial dimensions.
Since we are working with rather high resolutions ($768\times 768$) in combination with a small number of non-synthetical training data (747 samples), this compression has to be very strong to have a well-behaving diffusion process.
Wordstylist~\citep{nikolaidou2023wordstylist} applied a pre-trained model from Stable Diffusion for this task. 
Preliminary results showed that this does not scale to paragraphs
 (see \cref{fig:ae_reco} and \cref{tab:ae_reco}). 
For a $768\times 768$ input image, their compression method results in a feature matrix with shape $(4 \times 96 \times 96)$, where $4$ is the feature dimension and $96\times 96$ is the spatial dimension. 
We retrained the \ac{kl} regularized \ac{vae}~\citep{kingma2014vae,rezende2014vae} from LDM but with a smaller feature dimension in the latent.
To facilitate this increased compression rate, we extend the feedback of the paragraph-level reconstruction with pre-trained text recognition and writer-style models, leading to the latent shape of $(1 \times 96 \times 96)$.
This updates the overall loss term to:
\begin{equation}\label{eq:train}
    \mathcal{L}_{\text{EDS}} = \mathcal{L}_{\text{rec}} + w_{\text{KL}} \cdot \mathcal{L}_{\text{KL}} + 
    w_{\text{HTR}} \cdot \mathcal{L}_{\text{HTR}} + w_{\text{WI}} \cdot \mathcal{L}_{\text{WI}},
\end{equation}
where $\mathcal{L}_{\text{HTR}}$ denotes the loss term for text recognition, while $\mathcal{L}_{\text{WI}}$ incorporates the style task, scaled by $w_{\text{HTR}}$ and $w_{\text{WI}}$, respectively. To balance and better align the different loss terms given their varying value ranges, we introduce the weightings $w_{\text{KL}}$, $w_{\text{HTR}}$, and $w_{\text{WI}}$. $\mathcal{L}_{\text{rec}}$ is the applied $L_1$ reconstruction loss, and for regularization, the \ac{kl} divergence is applied, denoted as $\mathcal{L}_{\text{KL}}$, with the weighting $w_{\text{KL}}$ set to the default value $1 \cdot 10^{-6}$.
Note that we removed the discriminator loss due to unpredictable training behaviors.

The text recognizer, which is based on the approach by Kang et al.~\citeyearpar{kang2022pay}, is applied to full paragraphs to ensure readable and unmodified content.
The model combines a feature extractor, similar to $\mathcal{E}$ but not with shared weights, and a transformer model for encoding the features and producing the output predictions. 
Further, to apply the transformer model to the extracted features, we add adaptive two-dimensional positional encoding~\citep{lee20202dpe} to the feature encodings, similar to ~\citep{kang2022pay}. 
$\mathcal{L}_{\text{HTR}}$ is computed using cross-entropy, following sequence-to-sequence HTR approaches \citep{kang2022pay,wick2021bidirectional}.
The writer ID is used for correctly matching the writing style. 
We use a \ac{cnn} to predict the writer ID for increased data efficiency and reduced overfitting due to the sparsity of writing styles.\footnote{The appendix and code give a detailed view of the architectures of the different models.}

\subsection{Diffusion Model}{\label{sec:method:diffusion}}
Diffusion Models are generative models that learn a data distribution $p(x)$ to reverse a diffusion process. In this process, gradually added noise to the image results in normally distributed noise. It is parametrized by a Markov Chain of length $T$, which is the total number of time steps with values close to $T$, resulting in almost completely noised inputs.
Reversing that process in the Denoising U-Net from LDM~\citep{rombach2022ldm} $\epsilon_{\Theta}(z_t,t,c)$ is done by gradually removing the noise, where $z_t$ is the noised image at time step $t$ and $c$ stands for the conditioning described in \cref{sec:conditioning}. 
We added adaptive two-dimensional positional encoding~\citep{lee20202dpe} after the first projection layer of the spatial transformers for robust runs. 
This idea is inspired by \ac{htr} methods~\citep{kang2022pay}, which apply positional encoding to input images. We hypothesise that, while sharing the same style, different regions and words in the image are somewhat independent of each other. Preliminary experiments showed that without this extension, the model struggled to reconstruct entire paragraphs, instead producing a jigsaw-like arrangement of strokes and lines.
The objective of training the model parameters $\Theta$ is defined as
$\mathcal{L}_{\text{LDM}} = ||\epsilon - \epsilon_\Theta(z_t, t, c)||^2_{2}$.

\subsection{Conditioning}
\label{sec:conditioning}

We use conditioning as part of the architecture to prime the model with a specific style and a defined target text. 
Typically, just one modality is used as side input for the diffusion process model~\citep{yang2023dmsurvey}. In contrast, we fuse style and content with a transformer decoder for handwriting imitation.
The style encoder $\mathcal{E}_{\text{style}}$ consists of an initial convolutional layer, followed by 4 residual blocks with a total spatial downscaling factor of 128, followed by a final convolutional layer resulting in a spatial shape of ($6\times 6$). A small multi-layer perceptron head is used for pre-training $\mathcal{E}_{\text{style}}$ on writer classification. 

The encoded $x_{\text{text}}$ with added 1D sinusoidal positional encoding~\citep{vaswani2017transformer} and the embedded $x_{\text{style}}$ with added adaptive 2D positional encoding~\citep{lee20202dpe} are incorporated in a transformer model using cross-attention layers.
Cross-attention~\citep{vaswani2017transformer} employs regular multi-head attention, \ie,
    $
    \text{Att}(Q,K,V) = \text{softmax}(\frac{QK^T}{\sqrt{d}}) \cdot V \,,
    $
but with different inputs: $Q=W_Q\cdot \mathcal{E}_{\text{embed}}(x_{\text{text}})$, $K=W_K\cdot \mathcal{E}_{\text{style}}(x_{\text{style}})$, and $V=W_V\cdot \mathcal{E}_{\text{style}}(x_{\text{style}})$.
Note, for clarity, multi-head notation is omitted in the equations.

\subsection{Sampling}

\subsubsection{Classifier-free Guidance}

Another important part of diffusion models is the sampling for generating new latent representations, which, in our case, is based on additional conditioning. 
Recent developments have shifted towards adopting classifier-free guidance~\citep{ho2021classifierfree} over its predecessor, classifier guidance~\citep{dhariwal2021guidance}. This shift is not merely a matter of preference but is substantiated by empirical evidence suggesting enhanced performance in generating conditioned outputs that closely mimic the desired attributes. Our preliminary experiments in handwriting imitation verify this trend, indicating a superior fidelity in reproducing handwriting styles when utilizing classifier-free guidance. 

The basis for applying classifier-free guidance is a diffusion model which learns a conditional distribution $p(x|c)$ and an unconditional distribution $p(x)$ at the same time. We achieve this by replacing the conditioning information with an empty style image $x_\text{empty}$ and an empty string with a set probability $p=0.2$ during training. That allows us to strengthen the conditioning information during sampling by leveraging the scaled difference between the conditional and unconditional distribution. Mathematically classifier-free guidance equates to: 
\begin{equation}
    \epsilon_{t,c} = \epsilon_\Theta (x_t, t, c_{\text{empty}})+s\cdot (\epsilon_\Theta(x_t,t,c)-\epsilon_\Theta(x_t, t,c_{\text{empty}})),
\end{equation}
where $s$ is the scaling parameter and $c_{\text{empty}}$ is modeled as a blank page for style input and an empty string as target text. 
Here, $s$ controls the strength of the conditioning signal, where $s=0$ removes conditioning, $s=1$ applies it as given, and $s>1$ amplifies it, making the model adhere more strongly to the provided guidance.

\subsubsection{Ranked Resampling}
\label{sec:rerank}

Ding et al.~\citeyearpar{ding2023improving} improved the results by applying their progressive data filtering strategy. However, this technique only focuses on the character outputs and not on the style. They achieved the filtering by iteratively removing bad synthetic images below a certain confidence threshold and fine-tuning a pre-trained \ac{htr} to decide which samples to keep for the next round.
However, we focus our ranked resampling not only on legibility but also on style similarity.
Specifically, we generate $K$ samples from the same data point.
To compute style vectors for evaluating style similarity, we employ a traditional writer retrieval pipeline that involves local feature extraction followed by the computation of a global feature representation~\citep{Christlein15ICDAR,Christlein17PR,Christlein18DAS}. RootSIFT descriptors~\citep{Arandjelovic12} are extracted at SIFT keypoints~\citep{Lowe04} and subsequently jointly whitened and dimensionality-reduced through PCA~\citep{Christlein17PR}. The computation of the global feature representation is computed using multi-VLAD, where multiple VLAD encodings are once more PCA-whitened~\citep{Christlein15ICDAR}.
Style similarity is measured using cosine similarity between the style vector of the generated sample $\Tilde{x}$, and the target style vector $x_{\text{style}}$.
To measure readability, we use the \ac{cer} obtained from an \ac{htr} system trained exclusively on the training set paragraphs and additionally created synthetic images. The architecture of this system matches that of the \ac{htr} system in the encoder-decoder stage described in \cref{sec:methodology:encdec}.
Each sample is ranked based on these measures, allowing us to identify and select the samples that best balance stylistic fidelity with readability.
In the following, we define the ranking of the samples using $\text{rank}_{\text{WI}}$ for the style property and $\text{rank}_{\text{HTR}}$ for readability.

Since our ranked resampling approach generates $K$ samples per data point, the computational cost primarily arises from running the \ac{ldm} $K$ times, followed by inference using the \ac{htr} and \ac{wi} models. Given that $K$ remains relatively small (typically 1–10), we can mitigate the sorting of the data points, which should account for the term $K\log K$; the method scales linearly with the number of data points $N$, ensuring practical feasibility.

\section{Empirical Evaluation}
\label{sec:experiments}

\subsection{Dataset}

\subsubsection{IAM Handwriting Database}

The IAM database~\citep{marti2002iam} is used at the paragraph level.
For fine-tuning, we employ the 747 samples of the train split and the 116 samples of the validation split. Due to this low amount of training data, we created $50,000$ additional paragraphs with 365 true-type fonts from the internet and text from text generators.
We select a portion of the 336 IAM test paragraphs for testing to guarantee a writer-disjoint and, thus, zero-shot setting.
Therefore, we only use test samples of writers that do not appear in the train and validation sets and for which at least two samples are available. This criterion ensures that the priming information must stem from a different document. Consequently, we have assembled a collection of 247 documents authored by 72 writers.

\subsection{CVL Database}
For the out-of-distribution evaluation, we employ the CVL dataset~\cite{kleber2013cvl} at the paragraph level. It contains 1604 handwritten paragraphs across 310 unique writers in German and English. 
However, we had to remove any paragraphs containing an umlaut, as the IAM training alphabet does not contain these special characters. Thus, we ended up with 984 paragraphs, which we split into 108 for training, 31 for validation, and 845 for testing. From the original 310 writers, we assigned 22 for training, 282 for testing, and the remaining ones for validation.
This dataset is mainly used for \ac{wi}. In contrast, it is rarely applied for \ac{htr} because the training set is smaller than the test set.

\subsection{Metrics}
\subsubsection{Image Generation Quality via FID, KID, HWD} 

For natural images, the performance of generative models is commonly evaluated using \acf{fid}~\citep{heusel2017fid} and \acf{kid}~\citep{binkowski2018kid}. 
These metrics measure the similarity between real and generated images in a feature space extracted from a pre-trained deep neural network. Lower values indicate a closer resemblance between the generated and real samples, meaning improved realism and quality of the generated handwriting.
We evaluate them on paragraph and line levels.
Both metrics are tailored towards natural images with the underlying Inception model trained on ImageNet~\citep{deng2009imagenet}. However, the distribution of handwritten data is different. Therefore, Pippi et al.~\citeyearpar{pippi2023hwd} introduced a handwriting-specific line-based metric denoted as \ac{hwd},\footnote{HWD: \url{https://github.com/aimagelab/HWD}. Note: We slightly adapted the computation of the FID and KID so that the input is split into non-overlapping patches instead of using just the first square patch of the input.}
where a VGG16 backbone is trained on 100M rendered text lines and words to classify the calligraphic fonts. Similar to \ac{fid} and \ac{kid}, feature representations are finally used for comparing the distributions of different datasets. 
Lower \ac{hwd} values indicate that the generated samples better match the structural and stylistic properties of real handwriting.

\subsubsection{Style Assessment via Writer Identification} For assessing the stylistic accuracy, we rely on a learning-free \acf{wi} method.
The efficacy is then determined in a zero-shot setting, \ie, evaluating the test dataset in a leave-one-sample-out cross-validation where each sample is picked as query and the remaining samples are ranked according to their similarity to the query. From the ranks, Mean Average Precision (mAP) and top-1 accuracy are computed.
Higher mAP and top-1 values indicate that generated handwriting is more distinguishable as belonging to a specific writer, meaning better style preservation. A well-performing system should achieve high retrieval scores when querying generated samples against real samples of the same writer.
As \ac{wi} method, we follow the approach by Nikolaidou et al.~\citeyearpar{nikolaidou2023wordstylist} and rely on the same writer retrieval pipeline as outlined in \cref{sec:rerank}.

\subsubsection{Content Quality via HTR} Content preservation is commonly measured in terms of \acf{cer} with an \ac{htr} model comparing the target text with the generated text. The \ac{htr} model is trained on the genuine IAM training and test set.

\subsection{Implementation Details}

The experiments are focused on line and paragraph levels in the empirical evaluation because they are used in real-world scenarios.  
We compare against three state-of-the-art methods and use their implementation and pre-trained models for an unbiased evaluation: HiGAN+,\footnote{\url{https://github.com/ganji15/HiGANplus}} VATr,\footnote{\url{https://github.com/aimagelab/VATr}} and TS-GAN.\footnote{\url{https://github.com/herobd/handwriting_line_generation}} 
Note that VATr and HiGAN+ were mainly built for word-level handwriting generation and thus produce unrealistic text lines due to the stitching process.
For priming the style, we avoid using information from the same document. 
HiGAN+ needs just one word as style information, which is the lowest amount of all approaches. Therefore, a word image from the same writer's other document is sampled and used as style input. For VATr, 15-word images are sampled from the other document, while TS-GAN gets a random line image as style information. 
Our approach works on paragraphs, so our model uses a paragraph from the same writer's other document as style input. 
Please refer to the appendix for a detailed overview of the parameters and settings.

\subsection{Results}

\begin{figure}[t]
    \centering
    \includegraphics[width=\linewidth]{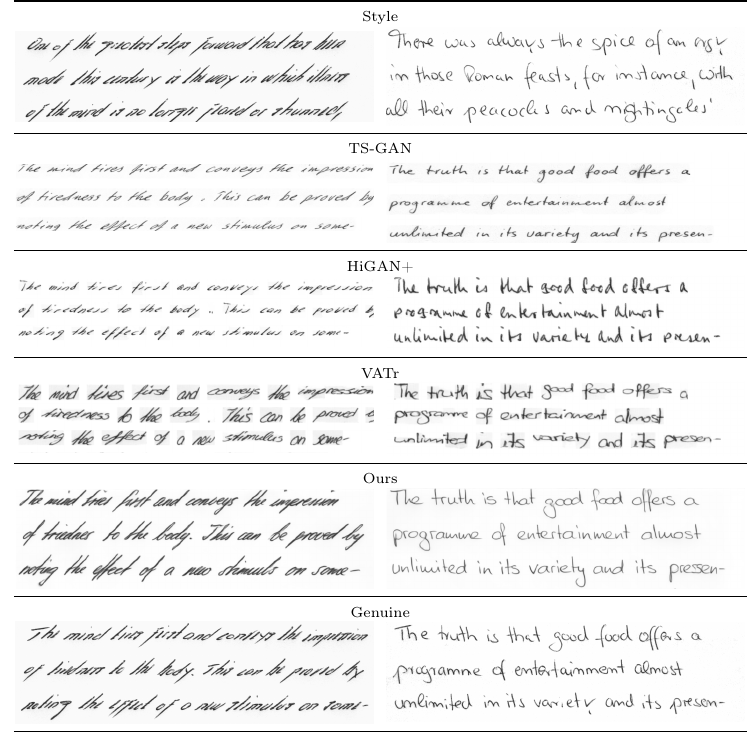}
    \caption{Comparison of text generation and style imitation performances based on a style (top) and target text of a genuine sample (bottom). 
    Images were sampled at random and cropped after the three lines.}
    \label{fig:quali}
\end{figure}

In this section, we analyse style and content preservation at the paragraph level. Additionally, we apply line segmentation~\citep{kodym2021pero} to assess our method at a more granular line level, addressing concerns about layout patterns versus intended content and style nuances. 
Finally, we conclude with ablation studies analysing the performance using synthetically generated data to fine-tune an \ac{htr} model, the generalisation capabilities on out-of-distribution data, and different parts of the framework.

\subsubsection{Qualitative Results}
In a qualitative analysis, we let the models write the same text of a given paragraph in a specific writing style. \Cref{fig:quali} showcases two samples picked at random.  
In contrast to state-of-the-art methods, our method shows a consistent writing style, which is closer to the given style input and thus also closer to the original genuine sample (s.\ bottom of \cref{fig:quali}). 
A general problem among all approaches seems to be the wrong selection of glyphs, which are still not close enough to the style sample.
Note that background artifacts around the paragraph and also around words stem from the IAM dataset. The model only reproduces this style pattern from the genuine data.

\begin{table}[t]
    \centering
    \caption{Paragraph-level style evaluation using \acf{wi} performance. Q and K stand for query and key, respectively. The top rows show the \ac{wi} performance on the IAM dataset and a stitched version. The stitching post-processing used for word-based imitation methods does not alter the \ac{wi} performance. We evaluate two modalities: using pure synthetic, \ie, generated samples, and a mix of synthetic and genuine samples. All results are given in [\%].   
    } 
    \setlength{\tabcolsep}{.55em}
    \begin{tabular}{lccccc}
        \toprule
         & \multicolumn{2}{c}{\textbf{Q: Synth \& K: Synth}} & & \multicolumn{2}{c}{\textbf{Q: Synth \& K: Genuine}} \\
         \cmidrule{2-3}
         \cmidrule{5-6}
         & \textbf{top-1} $\uparrow$ & \textbf{mAP} $\uparrow$ & & \textbf{top-1} $\uparrow$ & \textbf{mAP} $\uparrow$ \\
         \midrule
        IAM  & $97.57$  & $97.36$ & & $97.57$  & $97.36$ \\
        IAM stitched  & $97.17$  & $97.30$ & & $97.17$  & $97.00$   \\
        \midrule
        TS-GAN  & $26.68 $  & $17.61  $ & & $\phantom{9}5.34  $  & $11.42  $   \\
        HiGAN+  & $37.69 $  & $29.80$ & & $30.00 $  & $37.51 $   \\
        VATr  & $63.68  $  & $ 59.43  $ & & $\phantom{9}7.13  $  & $14.40  $   \\        
        \midrule
        Ours & $82.35 $ & $80.20  $ & &  $50.23 $ &   $54.73  $ \\
        Ours + $\text{rank}_{\text{HTR}}$ & $80.16$ & $81.08$ & & $50.61$ & $56.57$ \\
        Ours + $\text{rank}_{\text{WI}}$ & $84.21$ & $83.23$ & & $\textbf{56.28}$ & $\textbf{61.06}$ \\
        Ours + $\text{rank}_{\text{HTR+WI}}$ & $\textbf{86.64}$ & $\textbf{83.48}$ & & $54.66$ & $59.40$ \\
        \bottomrule
    \end{tabular}
    \label{tab:paragraph_style_eval_synth}
\end{table}

\begin{figure}
    \centering
    \fbox{\includegraphics[width=0.95\textwidth]{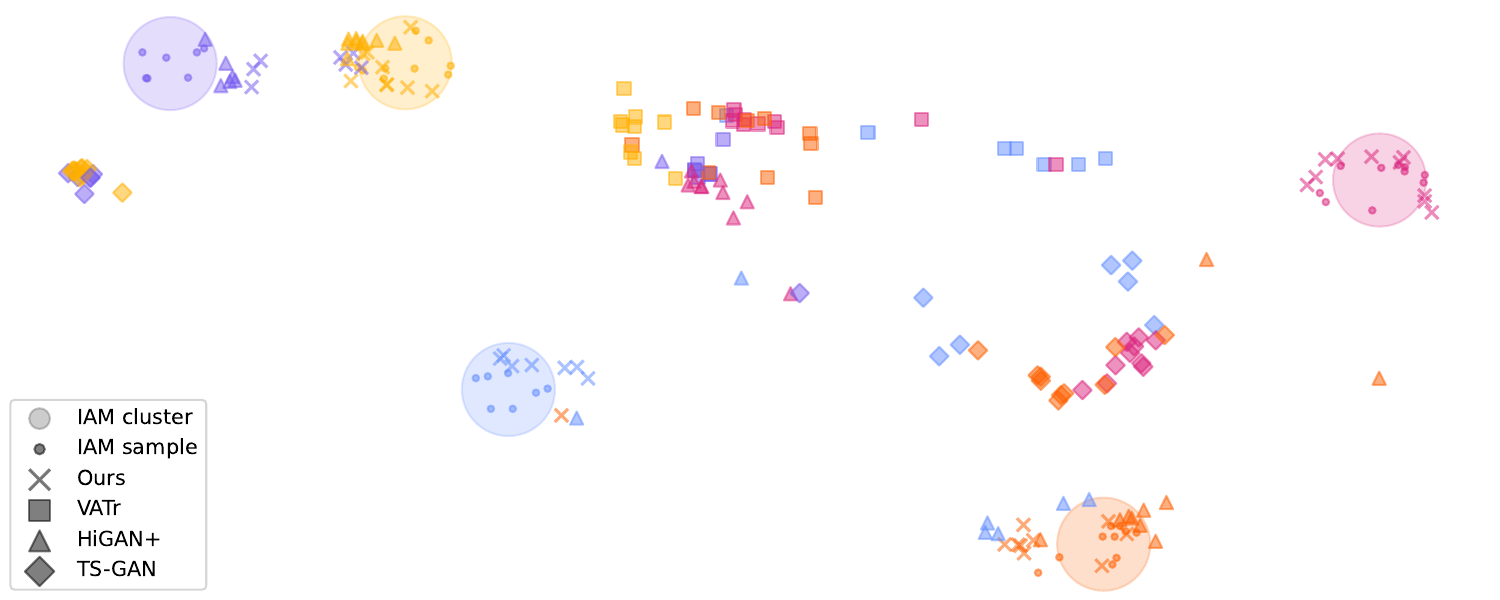}}
    \caption{UMap visualization of the five most present writers in the IAM test set, colour-coded in the plot. 
    It shows that our generated samples ($\times$) are much closer to the genuine samples ($\bullet$) than those generated by the other methods ($\blacksquare$, $\blacktriangle$, $\blacklozenge$).}
    \label{fig:umap}
\end{figure}

\subsubsection{Style Preservation}

\Cref{tab:paragraph_style_eval_synth} assesses style preservation on the paragraph level.
First, we evaluate the writer identification task exclusively on the imitated images (Query: Synth \& Key: Synth) to examine the consistency within the styles of the generated paragraphs.
Second, to verify the authenticity of the preserved genuine style, we treated the generated images as queries and calculated their \mbox{top-1} and mAP scores against the pool of real samples (Query: Synth \& Key: Genuine).
The top row shows the results on genuine IAM test data to validate the writer identification task as an evaluation metric. Additionally, we report the stitched IAM test data (IAM stitched) results to justify our stitching protocol, which was applied as post-processing for the comparison approaches.
The similarly high IAM and IAM stitched results highlight that (1)~our applied \ac{wi} method is effective and (2)~our stitching process does not influence the writer identification performance.

Our method significantly outperforms current state-of-the-art methods in both experiments (Q: Synth \& K: Synth and Q: Synth \& K: Genuine). 
Higher top-1 and mAP scores indicate that our generated handwriting is more stylistically consistent and closely matches the intended writer's style.
VATr~\citep{pippi2023archetypes} performs well for the synthetically generated images but cannot preserve the style of the given input style image. At the same time, HiGAN+~\citep{gan2022higanplus} performs similarly well in both experiments.
In addition to the baseline method, we evaluate the effect of different ranked resampling strategies. In particular, we rank the samples according to their performance in \ac{wi} ($\text{rank}_{\text{WI}}$), \ac{htr} ($\text{rank}_{\text{HTR}}$), or both ($\text{rank}_{\text{HTR+WI}}$). As expected, the results show that a ranked sampling using \ac{wi} is especially beneficial to preserve the input style (s.\ Synth+Genuine). The combination of \ac{wi} and \ac{htr} is slightly worse. 

\Cref{fig:umap} offers an intuitive visualization of our results, showcasing the distribution of documents from the five most prolific writers in our dataset. 
We applied UMap dimensionality reduction~\citep{mcinnes2020umap} to the L2-normalized global feature vectors obtained from the writer identification task.
In this plot, each writer is distinguished by a unique colour, and the cluster centers are depicted as large, transparent circles. 
Surrounding these central points, the genuine test data samples, represented by smaller dots, tend to cluster closely.
However, the representations generated by VATr~\citep{pippi2023archetypes} are mostly situated between the clusters of genuine writers, suggesting a less distinct association with any specific writer's style.
A similar observation can be made for TS-GAN~\citep{davis2020text}, for which the low-dimensional representations tend to mix across the clusters of genuine writers.
HiGAN+~\citep{gan2022higanplus} exhibits a somewhat better alignment in certain cases but struggles to accurately associate with the styles of the blue and red writers, indicating a partial success in style emulation. 
In contrast, the style vectors generated for our model's paragraphs demonstrate a notably closer affiliation with the intended writers' clusters, although with minor inaccuracies. For instance, a few blue samples are closer to the yellow cluster than their target blue centre.

\begin{table}[t]
    \centering
    \caption{Assessment of line-level and paragraph-level text generation. HWD, CER{\textsubscript{L}}, FID{\textsubscript{L}} and KID{\textsubscript{L}} are computed on the line level while FID{\textsubscript{P}} and KID{\textsubscript{P}} are calculated on the paragraph level. CER{\textsubscript{L}} is in [\%].  } 
    \setlength{\tabcolsep}{.55em}
    \begin{tabular}{lcccccc}
        \toprule
         &  \textbf{FID{\textsubscript{L}} }
          $ \downarrow $ & \textbf{KID{\textsubscript{L}}} $\downarrow$ & \textbf{FID{\textsubscript{P}}} $\downarrow$ & \textbf{KID{\textsubscript{P}}} $\downarrow$ & \textbf{HWD}$\downarrow$ & \textbf{CER{\textsubscript{L}}} $\downarrow$ \\
         \midrule
        TS-GAN & $84.56$ & $0.137 $ & $86.83$ & $0.081$ & $1.30 $ &$\textbf{0.80}$ \\
        HiGAN+ & $39.98$ & $ 0.050 $ & $81.65$& $0.086$  & $1.32 $ & $3.98$  \\
        VATr & $33.72$ & $0.043 $ & $71.32$ & $0.067$& $1.50 $ & $5.46$ \\
        \midrule
        Ours + $\text{rank}_{\text{HTR+WI}}$ & $\textbf{15.87}$ &  $ \textbf{0.018}$ & $\textbf{22.38}$ & $\textbf{0.010}$  & $\textbf{0.86}$ & $4.95$ \\
        \bottomrule
    \end{tabular}
    \label{tab:line_eval_reworked}
\end{table}

Style preservation on a line level draws different results for the comparison approaches, as seen in \cref{tab:line_eval_reworked}.
For \ac{fid} and \ac{kid}, VATr~\citep{pippi2023archetypes} outperforms the others, where TS-GAN~\citep{davis2020text} drastically drops in performance. Conversely, for \ac{hwd} metrics, HiGAN+~\citep{gan2022higanplus} and TS-GAN~\citep{davis2020text} exceed VATr's performance. 
Our approach in combination with reranking for style and content achieves by far the best scores including \ac{fid}, \ac{kid} and \ac{hwd} on both line (FID\textsubscript{L}, KID\textsubscript{L}, \ac{hwd}) and paragraph-level (FID\textsubscript{P}, KID\textsubscript{P}). 
Note that for \ac{hwd}, we used the provided line-level outputs from the other methods, while for our approach, we first applied line segmentation~\citep{kodym2021pero} to extract individual lines from the generated paragraphs before computing \ac{hwd}.

\subsubsection{Content Preservation}

Another crucial aspect of handwriting imitation is text preservation, which we assess using the \ac{cer}.
Table~\ref{tab:line_eval_reworked} shows \ac{cer} results at the line level, with lower \ac{cer} values indicating fewer transcription errors. 
While TS-GAN~\citep{davis2020text} achieves the lowest \ac{cer}, it does so at the expense of poor style preservation, illustrating the trade-off between content accuracy and visual fidelity. In contrast, our approach maintains a low \ac{cer} while significantly outperforming other methods in style preservation, demonstrating its ability to balance both aspects effectively.

We also evaluated our approach on paragraph level, resulting in a \ac{cer}{\textsubscript{P}} of $4.77\,\%$. However, problems arise when dealing with lines longer than 75 characters. There, the \ac{cer}{\textsubscript{P}} raises to about $30\,\%$. We argue that the \ac{htr} model cannot cope with the extreme downscaling of line images.

\subsection{Ablations Study}

\begin{table}[t]
    \centering
    \small
    \caption{CER results \lbrack \%\rbrack~evaluated on real CVL test data for HTR models fine-tuned on synthetically recreated CVL train data.} 
    \begin{tabular}{cccccc}
        \toprule
        && \multicolumn{4}{c}{\textbf{Synthetic $\text{CVL}_{\text{train}}$}} \\
        \cmidrule{3-6}
        W/o finetuning & Genuine & Ours & TS-GAN & HiGAN+ & VATr \\   
        \midrule
        $16.36$ & $\phantom{0}6.60$ & $\textbf{12.73}$ & $15.19$ & $14.48$ & $15.41$ \\
        \bottomrule
    \end{tabular}
    \label{tab:htr}
\end{table}

\begin{table}
    \centering
    \small
    \caption{Out-of-distribution style evaluation on $\text{CVL}_{\text{train}}$. Q and K stand for query and key, respectively. All results are given in [\%].}
    \begin{tabular}{lcccccc}
        \toprule
         & \multicolumn{2}{c}{\textbf{Q: Synth \& K: Synth}} & & \multicolumn{2}{c}{\textbf{Q: Synth \& K: Genuine}} \\
         \cmidrule{2-3}
         \cmidrule{5-6}
         & \textbf{top-1} $\uparrow$ & \textbf{mAP} $\uparrow$ & & \textbf{top-1} $\uparrow$ & \textbf{mAP} $\uparrow$ & \textbf{HWD} $\downarrow$\\
         \midrule
         CVL & $100.00$ & $99.92$ & & $100.00$ & $99.92$ & - \\
         \midrule
        TS-GAN  & $\phantom{0}6.48$  & $11.34$ & & $\phantom{0}7.41$  & $12.92$ & $1.43$  \\
        HiGAN+  & $39.81$  & $33.07$ & & $32.41$  & $35.94$ & $1.37$  \\
        VATr  & $30.56$  & $40.10$ & & $\phantom{0}8.33$  & $13.70$ & $1.45$  \\        
        \midrule
        Ours & $\textbf{68.52}$ & $\textbf{68.87}$ & &  $\textbf{48.15}$ &   $\textbf{48.01}$ & $\textbf{1.01}$ \\
        \bottomrule
    \end{tabular}
    \label{tab:cvl_wi_train}
\end{table}

\begin{table}[t]
    \centering
    \small
    \caption{Out-of-distribution style evaluation on $\text{CVL}_{\text{test}}$. Q and K stand for query and key, respectively. All results are given in [\%].} 
    \begin{tabular}{lcccccc}
        \toprule
         & \multicolumn{2}{c}{\textbf{Q: Synth \& K: Synth}} & & \multicolumn{2}{c}{\textbf{Q: Synth \& K: Genuine}} \\
         \cmidrule{2-3}
         \cmidrule{5-6}
         & \textbf{top-1} $\uparrow$ & \textbf{mAP} $\uparrow$ & & \textbf{top-1} $\uparrow$ & \textbf{mAP} $\uparrow$ & \textbf{HWD} $\downarrow$\\
         \midrule
         CVL & $98.58$ & $98.04$ & & $98.58$ & $98.04$ & - \\
         \midrule
        TS-GAN  & $\phantom{9}0.24$  & $\phantom{9}1.63$ & & $\phantom{9}0.83$  & $\phantom{9}2.18$ & $1.65$   \\
        HiGAN+  & $\phantom{9}4.85$  & $\phantom{9}7.24$ & & $\phantom{9}4.73$  & $\phantom{9}8.37$ & $1.37$  \\
        VATr  & $\phantom{9}2.37$  & $\phantom{9}4.96$ & & $\phantom{9}1.18$  & $\phantom{9}2.80$  & $1.62$ \\        
        \midrule
        Ours & $\textbf{13.96}$ & $\textbf{21.55}$ & &  $\textbf{11.24}$ &   $\textbf{17.11}$ & $\textbf{1.04}$ \\
        \bottomrule
    \end{tabular}
    \label{tab:cvl_wi_test}
\end{table}

\subsubsection{Synthetic Data for Handwritten Text Recognition}
One of the primary purposes of generative models is to use them for downstream tasks. Here, we evaluated the usefulness of different handwriting imitation approaches for creating synthetic data for training \ac{htr} systems. We used a pre-trained \ac{htr} model (synthetic fonts+real IAM train), fine-tuned it on synthetically generated CVL~\citep{kleber2013cvl} training data, and evaluated it on real $\text{CVL}_{\text{test}}$. The CVL dataset was chosen because it is challenging for \ac{htr}. 
\Cref{tab:htr} demonstrates that our approach surpasses current methods. However, there is still a gap between genuine and synthetic data, suggesting that the handwriting imitation task needs to be improved to replace or increase the amount of real data samples.

\subsubsection{Style Generalisation Capabilities with Out-of-Distribution Data}\label{sec:results:cvl}
We analysed how well the style is preserved on out-of-distribution data. We applied our IAM-trained models on CVL data, following the same evaluation protocol as for IAM.
\Cref{tab:cvl_wi_train} and \cref{tab:cvl_wi_test} show worse results on CVL than on IAM but still significantly better than other approaches. Additionally, top-1 and mAP are considerably worse on $\text{CVL}_{\text{test}}$ than on $\text{CVL}_{\text{train}}$. We hypothesize that the matching is much harder due to the larger test set and fewer samples per writer.

\begin{figure}[t]
    \centering
    \begin{subfigure}{0.47\textwidth}
        \fbox{\begin{tikzpicture} 
        \begin{scope}[spy using outlines={rectangle,lens={scale=6},height=0.8cm,width=4cm, connect spies, ultra thick}, node distance=1cm]
        \node[](genuine){%
	       \includegraphics[width=1\textwidth, trim={0cm 21cm 0cm 0cm}, clip]{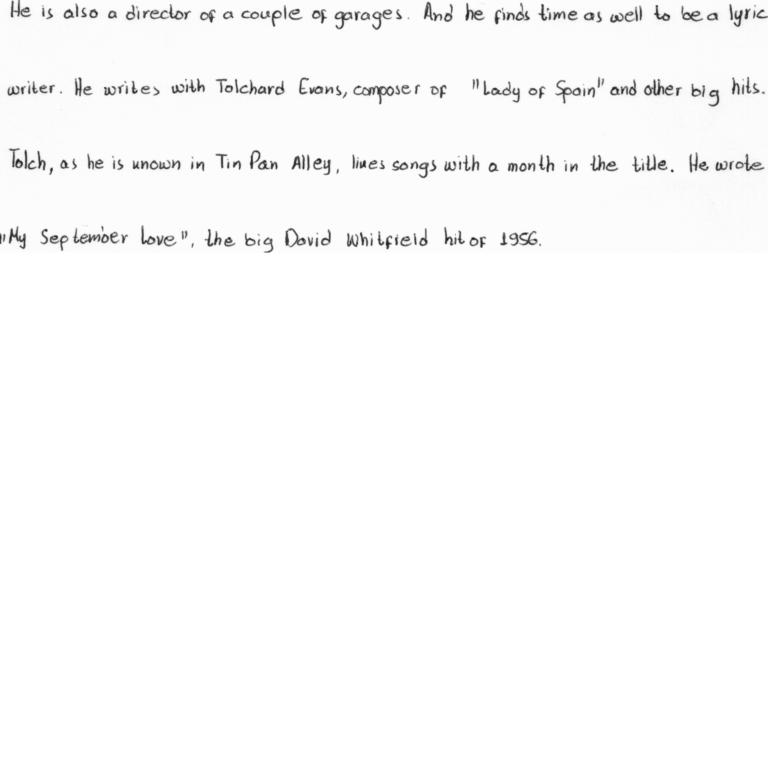}
        };
        \spy[orange,thick] on ($(genuine.north)+(-0.75cm,-0.22cm)$) in node (s1) [below=0.65cm of
            genuine.north];
        \end{scope}
        \end{tikzpicture}} 
        \caption{Genuine}
        \label{fig:ae_reco_a}
    \end{subfigure}
    \hfill
    \begin{subfigure}{0.47\textwidth}
        \fbox{\begin{tikzpicture} 
        \begin{scope}[spy using outlines={rectangle,lens={scale=6},height=0.8cm,width=4cm, connect spies, ultra thick}, node distance=1cm]
        \node[](kl_vanilla){%
	       \includegraphics[width=1\textwidth, trim={0cm 21cm 0cm 0cm}, clip]{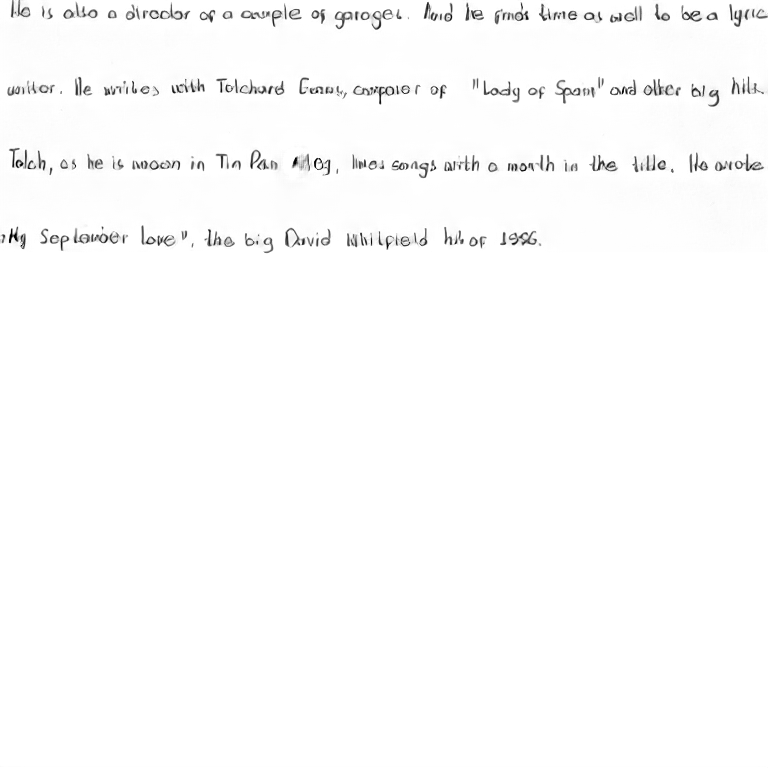}
        };
        \spy[orange,thick] on ($(kl_vanilla.north)+(-0.75cm,-0.22cm)$) in node (s3) [below=0.65cm of
            kl_vanilla.north];
        \end{scope}
        \end{tikzpicture}}
        \caption{Stable Diffusion VAE}
        \label{fig:ae_reco_b}
    \end{subfigure}
    \begin{subfigure}{0.47\textwidth}
        \fbox{\begin{tikzpicture} 
        \begin{scope}[spy using outlines={rectangle,lens={scale=6},height=0.8cm,width=4cm, connect spies, ultra thick}, node distance=1cm]
        \node[](kl_extra){%
	       \includegraphics[width=1\textwidth, trim={0cm 21cm 0cm 0cm}, clip]{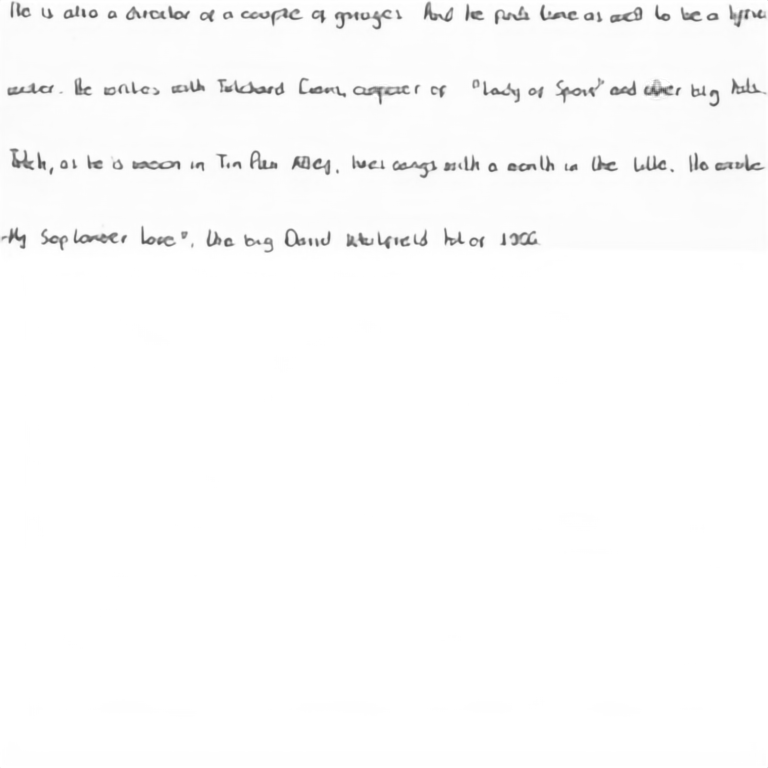}
        };
        \spy[orange,thick] on ($(kl_extra.north)+(-0.75cm,-0.22cm)$) in node (s4) [below=0.65cm of
            kl_extra.north];
        \end{scope}
        \end{tikzpicture}}
        \caption{Ours (vanilla)}
        \label{fig:ae_reco_c}
    \end{subfigure}    
    \hfill
    \begin{subfigure}{0.47\textwidth}
        \fbox{\begin{tikzpicture} 
        \begin{scope}[spy using outlines={rectangle,lens={scale=6},height=0.8cm,width=4cm, connect spies, ultra thick}, node distance=1cm]
        \node[](stable){%
	       \includegraphics[width=1\textwidth, trim={0cm 21cm 0cm 0cm}, clip]{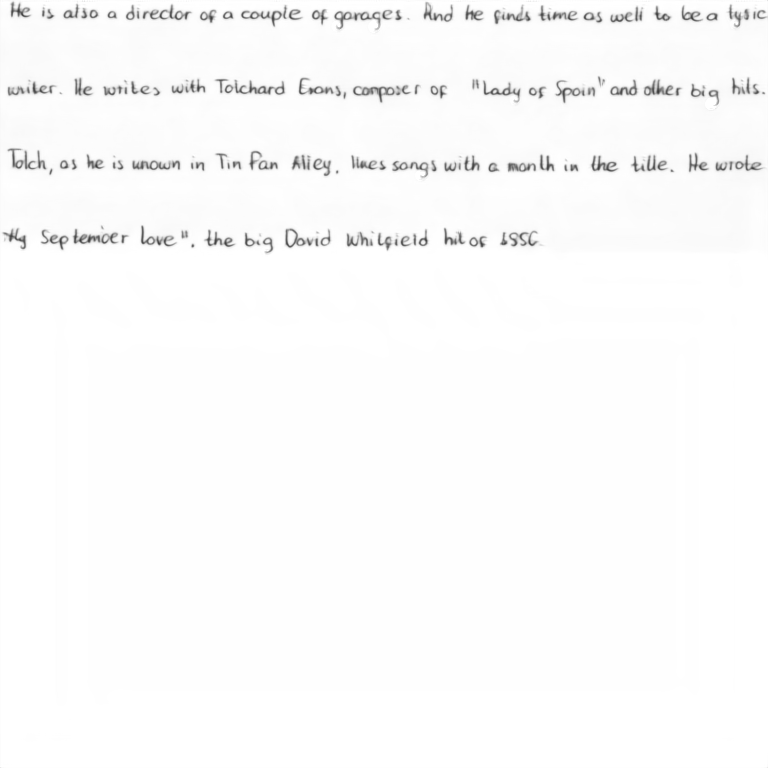}
        };
        \spy[orange,thick] on ($(stable.north)+(-0.75cm,-0.22cm)$) in node (s2) [below=0.65cm of
            stable.north];
        \end{scope}
        \end{tikzpicture}}
        \caption{Ours (w/ HTR and WI)}
        \label{fig:ae_reco_d}
    \end{subfigure}
    \caption{Qualitative comparison of the paragraph reconstructions showing that the additional \ac{htr} and \ac{wi} losses are beneficial. }
    \label{fig:ae_reco}
\end{figure}

\begin{table}[t]
    \centering
    \caption{Assessment of encoder-decoder stage's reconstruction performance. The \ac{htr} results are produced by paragraph-based (CER{\textsubscript{P}}) and line-based (CER{\textsubscript{L}}) \ac{htr} models. All results are in [\%].}
    \setlength{\tabcolsep}{.55em}
    \begin{tabular}{lcccc}
    \toprule
     & \textbf{MAE} $\downarrow$ & \textbf{MSE} $\downarrow$ & \textbf{CER{\textsubscript{P}}} $\downarrow$ & \textbf{CER{\textsubscript{L}}} $\downarrow$ \\
    \midrule
    Genuine & $0.00$ & $0.00$ & $\phantom{1}1.05$ & $1.42$ \\
    \midrule
    StableDiffusion & $1.32$ & $\textbf{0.13}$ & $\phantom{1}5.27$ & $3.39$ \\
    Ours (vanilla) & $1.27$ & $0.15$ & $14.50$ & $9.95$ \\
    Ours (HTR+WI loss) & $\textbf{1.25}$ & $0.14$ & $\textbf{\phantom{1}3.19}$ & \textbf{2.29} \\
    \bottomrule
    \end{tabular}
    \label{tab:ae_reco}
\end{table}

\subsubsection{Encoder-Decoder Capabilities}

Wordstylist~\citep{nikolaidou2023wordstylist} demonstrated great success when applying LDMs for word-level handwritten text generation. They utilized pre-trained weights from Stable Diffusion~\citep{rombach2022ldm}~\footnote{\url{https://huggingface.co/runwayml/stable-diffusion-v1-5}} for their encoder-decoder stage.
We evaluate this approach and compare it to ours in \cref{tab:ae_reco}.
The reconstruction quality is first assessed using the standard metrics \ac{mae} and \ac{mse}. While these metrics show satisfactory numerical results, their practical significance for handwritten text reconstruction is limited. In traditional image processing tasks, small \ac{mae} and \ac{mse} differences indicate better pixel-wise similarity. However, in the context of handwriting, even minor pixel variations can significantly impact the legibility and stylistic fidelity of the text.
To further investigate this, we apply both paragraph-based and line-based \ac{htr} models to the reconstructed samples.
The models were trained on IAM's train and test data to decipher the different writing styles. 
The results reveal a significant increase in CER{\textsubscript{P}} (paragraph-based) and CER{\textsubscript{L}} (line-based) when using Stable Diffusion. 
When using handwritten paragraphs for direct training of a \ac{vae} from scratch with default loss terms,
 we observe even higher \ac{cer} values.
By contrast, the performance is improved when integrating the proposed writer and handwritten text recognition losses into training.
This is supported by a qualitative analysis of \cref{fig:ae_reco}, where \subref{fig:ae_reco_a} shows the original input to the encoder-decoder stage. %
Among the reconstructions without latent space modifications, \subref{fig:ae_reco_d} shows the closest resemblance to the original image despite a slight blurriness. The reconstruction from Stable Diffusion \subref{fig:ae_reco_b} alters certain characters, such as transforming the ``cou'' in ``couple'' into characters that more closely resemble ``au'', making them challenging to read. This observation aligns with the quantitative findings, where the default VAE exhibits reconstructions with significantly reduced readability, mirroring the high CER values.

\begin{table}[t] 
    \centering
    \caption{Comparison of different variations. Q and K stand for query and key, respectively. In ``Ours + Cosine'' a cosine scheduler is used instead of a linear scheduler while in ``Ours + No $'\backslash n'$'' new line tokens are removed in the target text. All results are computed with the best ranked samples based on $\text{rank}_{\text{WI}}$ and $\text{rank}_{\text{HTR}}$. Results are given in [\%].}
    \setlength{\tabcolsep}{.55em}
    \begin{tabular}{lccccccc}
        \toprule
         & \multicolumn{2}{c}{\textbf{Q: Synth \& K: Synth}} & & \multicolumn{2}{c}{\textbf{Q: Synth \& K: Genuine}}  \\
         \cmidrule{2-3}
         \cmidrule{5-6}
         & \textbf{top-1} $\uparrow$  & \textbf{mAP}  $\uparrow$ & & \textbf{top-1} $\uparrow$  & \textbf{mAP}  $\uparrow$  & & \textbf{\ac{cer}{\textsubscript{L}}} $\downarrow$ \\
         \midrule
        Ours                        & $\textbf{86.64}$ & $\textbf{83.48}$ & & $54.66$ & $59.40$ & & $\textbf{\phantom{3}4.95}$ \\
        Ours + Cosine  & $84.62$ & $81.59$ & & $\textbf{60.73}$ & $64.21$ & & $\phantom{3}7.03$  \\
        Ours + No $'\backslash n'$  & $80.57$ & $ 77.55$ & & $59.11$ & $\textbf{66.44}$ & & $34.84$ \\
        \bottomrule
    \end{tabular}
    \label{tab:variations}
\end{table}

\subsubsection{Cosine Scheduler and New-Line-Token-Free Variants}

We investigated two alternative versions of our approach: one employing a cosine scheduler~\citep{nichol2021cosine} to prioritize the general layout of text, and another omitting new line tokens, leaving the model to determine line initiations autonomously. 
As shown in \cref{tab:variations}, both modifications exhibit a similar intrinsic synthetic style but outperform our main approach in reflecting the handwriting styles in the genuine data.
However, these approaches come with a trade-off in terms of legibility.
Employing the cosine scheduler increases the \ac{cer}{\textsubscript{L}} to just over $7\%$, and removing newline tokens leads to a \ac{cer}{\textsubscript{L}} nearing $35\,\%$.
Additionally, qualitative assessments of the newline token-free variant revealed tendencies of the model to duplicate or omit words.

\begin{figure}[t]
	\centering
	\includegraphics[]{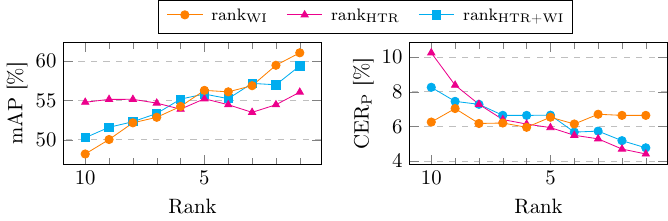}
\caption{Evaluating the effect of different ranking methods on the style consistency (mAP) and content preservation (\ac{cer}{\textsubscript{P}}). Ranking simultaneously by writer identification (WI) and by handwritten text recognition (HTR) provides a good compromise.}
\label{fig:ranking}

\end{figure}

\subsubsection{Ranking Effect}

In \cref{fig:ranking}, we analyse the impact of different ranking strategies on the performance of our baseline method on writer identification (mAP) and (paragraph-based) \ac{htr} (\ac{cer}{\textsubscript{P}}). 
The genuine writer style (\cref{fig:ranking} left) is best preserved when using only \ac{wi} feedback for ranking ($\text{rank}_{\text{WI}}$), achieving an mAP of above $60\,\%$ for the top-ranked samples (rank 1). It is unaffected by \ac{htr} feedback ($\text{rank}_{\text{HTR}}$), which remains stable at a mean performance level of approximately $55\,\%$ mAP. The combined ranking ($\text{rank}_{\text{HTR+WI}}$) positively impacts style outputs, though less effectively than $\text{rank}_{\text{WI}}$, reaching an mAP of nearly $60\,\%$ for the top rank.

Regarding content preservation (\cref{fig:ranking} right), applying $\text{rank}_{\text{HTR}}$ notably improves the outcomes, reducing the \ac{cer} to approximately $4\,\%$ for the top rank. $\text{rank}_{\text{HTR+WI}}$ also lowers the \ac{cer}, though not as effectively as $\text{rank}_{\text{HTR}}$, yet still achieving comparable results. 
In contrast, $\text{rank}_{\text{WI}}$ maintains a \ac{cer} between $6\,\%$ and $7\,\%$ across the different ranks.
It is important to highlight that implementing a ranking strategy utilizing both HTR and WI feedback results in a significant improvement in mAP, approximately five percentage points above the mean, and a concurrent enhancement in \ac{cer}{\textsubscript{P}}, approximately two percentage points better than the mean. Thus, this strategy strikes a meaningful balance between style and content preservation.

\section{Discussion}
\label{sec:discussion}

\Cref{tab:paragraph_style_eval_synth} shows that the writer style characteristics are well preserved, especially for synthetic samples. But even when imitating real handwriting captured on genuine images, the model shows realistic results. 
This is further demonstrated in the UMap plot (\cref{fig:umap}) where our method produces samples much closer to the original ones. 
Although our method achieves excellent replication of the desired style, the target text occasionally contains duplicate or incorrectly swapped characters, a flaw not seen with alternative methods. However, the \ac{cer}{\textsubscript{L}} reported in \cref{tab:line_eval_reworked} might not accurately represent the true \ac{cer}. 
We computed a $2.29\,\%$ \ac{cer}{\textsubscript{L}} on paragraphs reconstructed from the latent representations derived from the original images, ultimately setting a rather high baseline.
In the reconstruction quality results (\cref{tab:ae_reco}), readability is lower compared to genuine data processed with an omniscient \ac{htr} model. This sets the lower boundary for readability. 

\subsubsection*{Limitations with Out-of-Distribution Data}
Challenges with out-of-distribution data primarily arise from two sources: the target text and the style image.
For the target text, there is a small bias towards known words, which stems from a limited diversity in the paragraph training data. While large diffusion models are typically trained on millions of unique images, our training involved only $747$ real images and generated $\approx\!4000$ unique lines, which we permuted and stitched into a total of $50,000$ synthetic paragraphs, where every image contained 3 to 13 lines and 5 to 101 characters per line.
Additionally, the results degrade when the target text takes on uncommon paragraph forms, such as paragraphs containing only a single word or paragraphs with long lines exceeding 101 characters per line. We believe this issue arises because the current KL-regularized latent space representation does not adequately capture the overall rigid structure of handwritten paragraphs. As a result, the network struggles to generalize to rare out-of-distribution samples. This could be addressed by incorporating prior knowledge about the semantic structure of handwritten paragraphs into the regularization of the latent space. Alternatively, a simpler solution could involve incorporating additional synthetic and stitched real training data, with a particular focus on these edge cases.
Furthermore, stitching real data can introduce new artifacts, such as overly regular layouts.
For the style image, the distribution of out-of-distribution styles must align closely with the training data, particularly for real-world applications.
The results in \cref{sec:results:cvl} show the adaptation of the different methods to this case. Here, we can see that the results moderately decrease for $\text{CVL}_{\text{train}}$. We hypothesize that this could be due to the cleaner nature of the CVL data compared to the IAM data, which contains some artifacts, such as background gradients.
The biggest drop is with $\text{CVL}_{\text{test}}$, which could stem in addition from the fact that this dataset split has a big pool of many unseen writers ($283$) on $845$ paragraphs, making good top-1 and mAP results more challenging. This hypothesis is supported by the fact that the HWD stayed mostly consistent for our approach between $\text{CVL}_{\text{train}}$ and $\text{CVL}_{\text{test}}$.

\subsubsection*{Compute Time Comparison}
Computational efficiency is a key factor when applying generative models in real-world scenarios. To evaluate inference speed, we measured the time required to generate CVL paragraphs on an NVIDIA A40 GPU.
Our results confirm that GAN-based approaches significantly outperform diffusion-based models in speed. Among the tested methods, HiGAN+ is the fastest, generating a paragraph in 0.13 seconds, followed by TS-GAN (0.27 seconds) and VATr (0.28 seconds).
However, when comparing our method to another diffusion-based approach, WordStylist~\cite{nikolaidou2023wordstylist}\footnote{\url{https://github.com/koninik/WordStylist}}, we observe a substantial efficiency gain. WordStylist relies solely on the writer ID and does not need to extract the style from the image, requiring even fewer computations. Despite this, it still requires 13.44 minutes per paragraph due to its word-by-word generation process and a high number of sampling steps (600). In contrast, our approach reduces inference time to just 9.06 seconds per paragraph, achieving an average speed-up of 91, making it a more viable option for practical applications.

\subsubsection*{Possible Negative Implications}
A more appealing and realistic imitation of handwritten text poses several risks, particularly in forgery of sensitive documents such as wills, contracts, or historical records. Beyond document fraud, such a model could be exploited for identity theft, or falsification of handwritten evidence. These risks underscore the importance of robust forensic tools to detect AI-generated handwriting.

To counteract this, we make our approach and code publicly available to enable building countermeasures for these types of forgeries. There are already some initial works in this direction \citep{carriere2023detection}.
Further efforts in watermarking, authentication protocols, and forensic handwriting analysis could enhance security.

\section{Conclusion and Future Work}
\label{sec:conclusion}
In this study, we introduce a method that is capable of producing realistic-looking and style-consistent handwritten paragraphs in unseen writing styles. The approach is based on a refined latent diffusion model. By incorporating additional loss terms during the encoder-decoder phase, we achieved notable enhancements in both reconstruction quality and compression efficiency. Furthermore, the integration of style features with text embeddings proved to be effective for conditioning the denoising U-Net, demonstrating a successful application of our approach. 
Additionally, by imitating handwriting at the paragraph level rather than word by word, we significantly improved generation speed, making our method more efficient for practical applications.
Overall, our contributions not only advance the field of handwriting imitation but also hold the potential to benefit other document analysis tasks, particularly in scenarios characterized by limited data availability. 

Looking ahead, several opportunities exist to further enhance the model’s performance and applicability. 
Future work should focus on improving the encoder-decoder stage.
We hypothesise that a more compressed and structured latent space could enhance the generalisability and sampling speed of the diffusion model. Additionally, optimising this stage may help reduce artefacts, such as low-frequency gradients in the background.
Another key direction is increasing the amount of training data, which is crucial for fully leveraging diffusion models and mitigating issues related to out-of-distribution data, as discussed in \cref{sec:discussion}.
To bridge the computational gap between diffusion models and GANs, future research should explore reducing the number of sampling steps in combination with a more compressed latent space while maintaining high-quality output.

\backmatter

\bmhead{Acknowledgements and Funding}
\begin{itemize}
\item The authors gratefully acknowledge the scientific support and HPC resources provided by the Erlangen National High Performance Computing Center (NHR@FAU) of the Friedrich-Alexander-Universität Erlangen-Nürnberg (FAU). The hardware is funded by the German Research Foundation (DFG).
\item We acknowledge funding by the Deutsche Forschungsgemeinschaft (DFG, German Research Foundation) -- 416910787. 
\end{itemize}

\section*{Declarations}

\bmhead{Conflict of Interest}
The authors declare that they have no conflict of interest.

\bmhead{Data Availability}
This work does not propose new data. 

\bmhead{Code Availability}
The code can be accessed via GitHub link. %

\bmhead{Authors' Contributions}
All authors contributed to the study conception and design. Martin Mayr and Marcel Dreier performed the study analysis. Martin Mayr wrote the first draft of the manuscript, and all authors commented on previous versions of the manuscript. All authors read and approved the final manuscript.

\bigskip

\begin{appendices}

\section{Implementation Details}{\label{sec:implementation}}

\begin{table}[]
    \centering
    \scriptsize
    \caption{Hyperparamters of the VAEs}
    \begin{tabular}{lcc}
        \toprule
         & VAE w/ extra losses & Vanilla VAE \\
         \midrule
        Channels & 32 & 32 \\
        Dropout & 0.1 & 0.1 \\
        Channel Multiplier & 1,2,4,8 & 1,2,4,8 \\
        $z$ Shape & (1,96,96) & (1,96,96) \\
        Batch Size & 32 & 32 \\
        Optimizer & Adam & Adam \\
        Learning Rate & $1\cdot10^{-4}$ & $1\cdot10^{-4}$ \\
        
        \midrule
        Loss \\
        $w_{\text{KL}}$ & $1\cdot 10^{-6}$ & $1\cdot 10^{-6}$ \\
        $w_{\text{HTR}}$ & 0.3 & 0.0 \\
        $w_{\text{WI}}$ & 0.005 & 0.0 \\
        \midrule
        Augmentation \\
        p Dilation & 0.3 & 0.3 \\
        p Erosion & 0.3 & 0.3 \\
        p Distort with Noise & 0.3 & 0.3 \\
        \bottomrule
    \end{tabular}
    \label{tab:vae_params}
\end{table}

\begin{table}[] 
    \centering 
    \scriptsize
    \caption{Hyperparameters of the Paragraph HTR and Writer CNN systems.} \begin{tabular}{lcc} 
        \toprule 
        & Paragraph HTR & Writer CNN \\ 
        \midrule 
        Channels & 16 & 16 \\ 
        Channel Multiplier & 1,2,4,8 & 1,2,4,8,8 \\ 
        Dropout & 0.1 & 0.1 \\ 
        Hidden Size & 128 & 256 \\ 
        Optimizer & RAdam & Adam \\ 
        lr & - & 0.0001 \\ 
        $z$-shape-style & - & (6,6) \\ 
        Label Smoothing & 0.4 & 0.1 \\ 
        \# of Encoder Layers & 2 & - \\ 
        \# of Decoder Layers & 4 & - \\ 
        \# of Synthetic Samples & 3000 & 3000 \\ 
        \# of Real Samples & 747 & 747 \\ 
        \midrule 
        Augmentations & & \\ 
        p(Dilation) & 0.3 & 0.3 \\ 
        p(Erosion) & 0.3 & 0.3 \\ 
        p(Distort with Noise) & 0.3 & 0.3 \\ 
        p(Elastic Transform) & 0.3 & 0.3 \\
        p(Perspective Transform) & 0.3 & 0.3 \\ 
        p(Noisy Teacher) & 0.3 & 0.3 \\ 
        \bottomrule 
    \end{tabular} 
    \label{tab:htr_wi_params}
\end{table}

\subsection{Hyperparameters}

Hyperparameters for the first stage model are given in \cref{tab:vae_params}. The default VAE is defined as ``Vanilla VAE'', which is used in LDMs~\citep{rombach2022ldm}. $32$ channels are the initial amount in the feature dimension. The ``channel multiplier'' denotes the feature scaling from the outer to the inner blocks, each consisting of two ResnetBlocks. The encoder and decoder are mirrored.
The shape in the latent space is $(1,96,96)$, where $1$ is the feature dimension and $(96,96)$ is the spatial dimension. 
We used dilation, erasion, and distortion in combination with noise with a probability of $0.3$, each. Note that erosion cannot be applied when dilation is applied and vice versa.
``VAE w/ extra losses'' is the VAE model utilizing a Handwritten Text Recognition loss and Writer Identification loss.
In preliminary experiments, $w_{\text{HTR}}=0.3$ and $w_{\text{WI}}=0.005$ yielded the best results.

\Cref{tab:htr_wi_params} displays the hyperparameters for the HTR system for paragraphs. Similarly to the VAE encoder, the feature extractor consists of ResnetBlocks. The initial channel size of $16$ scales in the final block to $128$. $128$ also represents the model dimension (hidden size) in the transformer module. The Transformer consists of two encoder and four decoder layers. 
We added $3,000$ synthetic samples to the training process and applied label smoothing of $0.4$ to prevent overfitting.
Further heavy augmentations are utilized.

\begin{table}[]
    \centering
    \footnotesize
    \caption{Hyperparameters of the diffusion models.}
    \begin{tabular}{lccc}
        \toprule
         & Default & w/o NL token & Cosine scheduler \\
         \midrule
         $z$-shape & (1,96,96) & (1,96,96) & (1,96,96) \\
         Diffusion Steps & 1000 & 1000 & 1000 \\
         Noise Schedule & linear & linear & cosine \\
         Label Dropout & 0.2 & 0.2 & 0.2 \\
         Batch Size & 64 & 64 & 64  \\
         Iterations & 70k + 8k finetuning & 70k + 8k finetuning & 70k + 8k finetuning \\
         Warmup Steps & 10k & 10k & 10k \\
         Learning Rate & $5\cdot10^{-5}$ & $5\cdot10^{-5}$ & $5\cdot10^{-5}$ \\
         \midrule
         Denoising U-Net \\
         Channels & 256 & 256 & 256 \\
         Channel Multiplier & 1,2,4 & 1,2,4 & 1,2,4 \\
         Number of Heads & 1 & 1 & 1 \\
         \midrule
         Conditioning   \\
         Channels & 256 & 256 & 256 \\
         Dropout & 0.1 & 0.1 & 0.1 \\
         Context dimension & 1024 & 1024 & 1024 \\
         \# of Transformer Decoder Layers & 4 & 4 & 4 \\
         \midrule
         Augmentation \\
         p Gaussian Noise & 0.2 & 0.2 & 0.2 \\
         p Contrast & 0.2 & 0.2 & 0.2 \\
         p Brightness & 0.2 & 0.2 & 0.2 \\         
        \bottomrule
    \end{tabular}
    \label{tab:diff_params}
\end{table}

Hyperparameters of the VAE's writer identification model are described in \cref{tab:htr_wi_params}.
Like the VAE and HTR models, it uses ResnetBlocks but with increased downsampling resulting in a spatial shape of $(6,6)$. 
We also used $3,000$ synthetic samples and label smoothing in combination with heavy augmentations. When evaluating this model, we achieved an accuracy (top-1) of $90\%$.

To speed up the VAE, HTR, and WI models' training, we first trained them for 400 epochs on one- and two-line paragraphs, which provided good initial values for the main training process on paragraphs.

\Cref{tab:diff_params} shows the hyperparameters used for our default diffusion model, the new-line-free token approach, and the one with a cosine scheduler~\citep{nichol2021cosine} instead of a linear scheduler.
We set the number of diffusion steps $T$ to $1000$. The models are pre-trained for 70k iterations on synthetic and real data. Afterwards, for 8k iterations, the model is fine-tuned only on the real samples. We applied similar values to our denoising U-Net as in LDM~\citep{rombach2022ldm}.
Gaussian noise, contrast, and brightness augmentations are used for improved results.

\subsection{Model Architectures}{\label{sec:architecture}}

In this section we show the architectural changes we conducted. 

\begin{figure}[h]
    \centering
    \begin{subfigure}[b]{0.95\textwidth}
        \centering
        \fbox{\includegraphics[height=4cm]{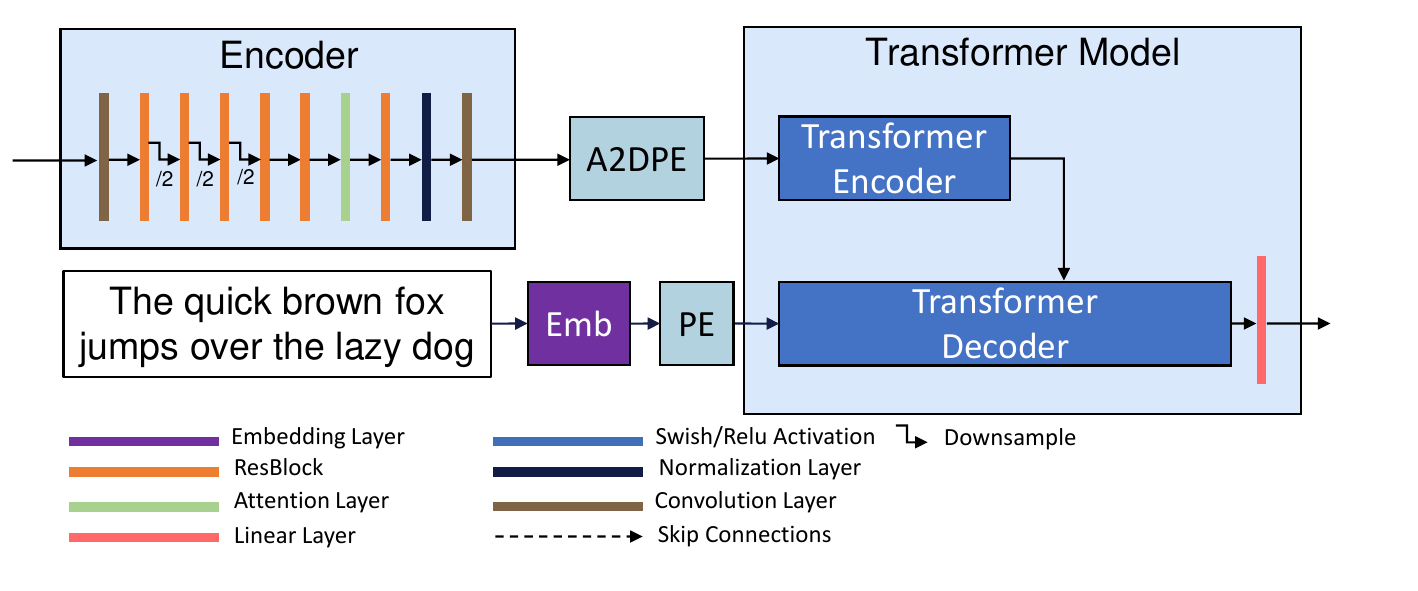}}
        \caption{Overview of Handwritten Text Recognition model.}
        \label{fig:htr}
    \end{subfigure}
    \hfill    
    \begin{subfigure}[b]{0.3\textwidth}
        \centering
        \fbox{\includegraphics[height=4.4cm]{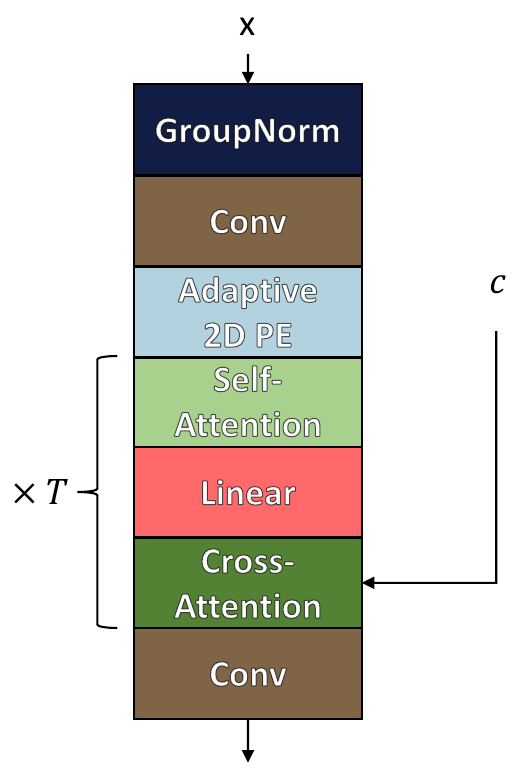}}
        \caption{Transformer block with adaptive 2D PE.}
        \label{fig:our_transformer_block}
    \end{subfigure}\hfill
    \begin{subfigure}[b]{0.6\textwidth}
        \centering
        \fbox{\includegraphics[height=4.4cm]{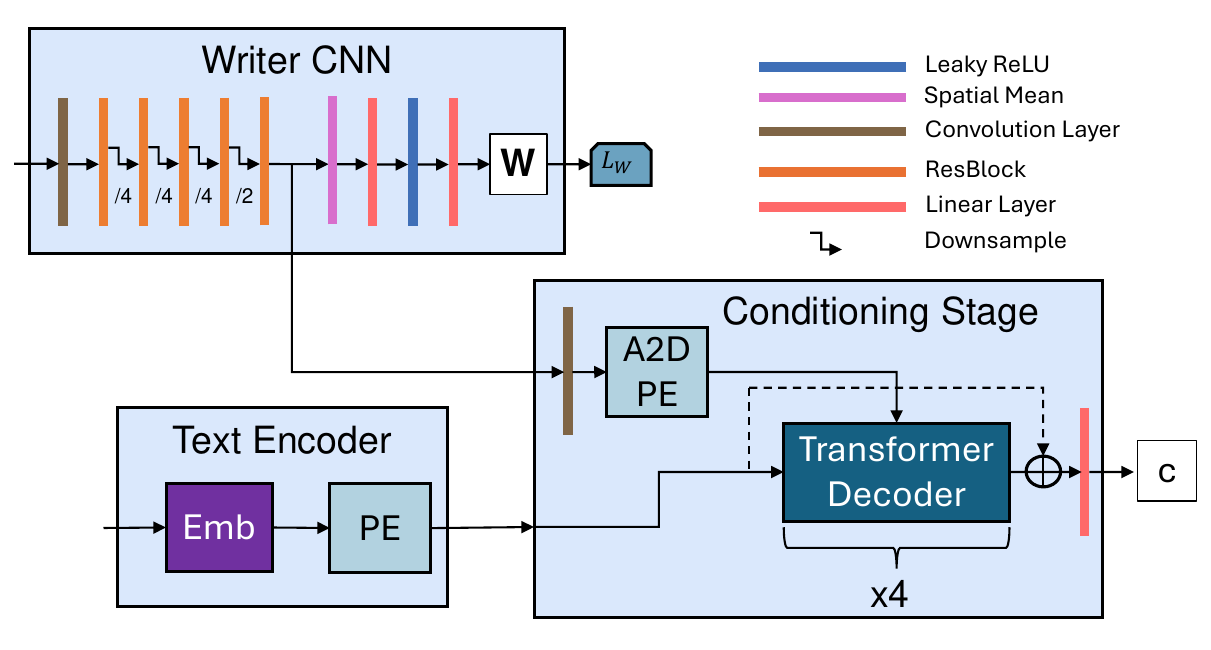}}
        \caption{Fusing writer information and text embeddings in the conditional stage.}
        \label{fig:conditioning_stage}
    \end{subfigure}
    
    \caption{Overview of the Handwritten Text Recognition model, the simplified transformer block, and the conditional stage of the model.}
    \label{fig:combined_figures}
\end{figure}

\Cref{fig:htr} shows the structure of the HTR model. It has a similar architecture as the HTR by Kang et al.~\citeyearpar{kang2022pay}. The main difference is the encoder which is resembled from ResBlocks.

Further, we extended the transformer blocks in the denoising U-Net with adaptive 2D positional encoding, see \cref{fig:our_transformer_block}. This is necessary to give the model an overall understanding of a paragraph.

\Cref{fig:conditioning_stage} displays the conditioning module. It consists of a Writer CNN and Text Encoder. The former one is pre-trained on writer labels. We use the latent representation after the last ResBlock added with adaptive 2D positional encoding~\citep{lee20202dpe} as keys and values for the cross-attention block in the Conditioning Stage.
Queries are the positional encoded text embeddings.

\section{Additional Results}{\label{sec:results}}

\begin{figure}[hp]
    \centering
    \begin{subfigure}{0.4\textwidth}
        \fbox{\includegraphics[width=\textwidth]{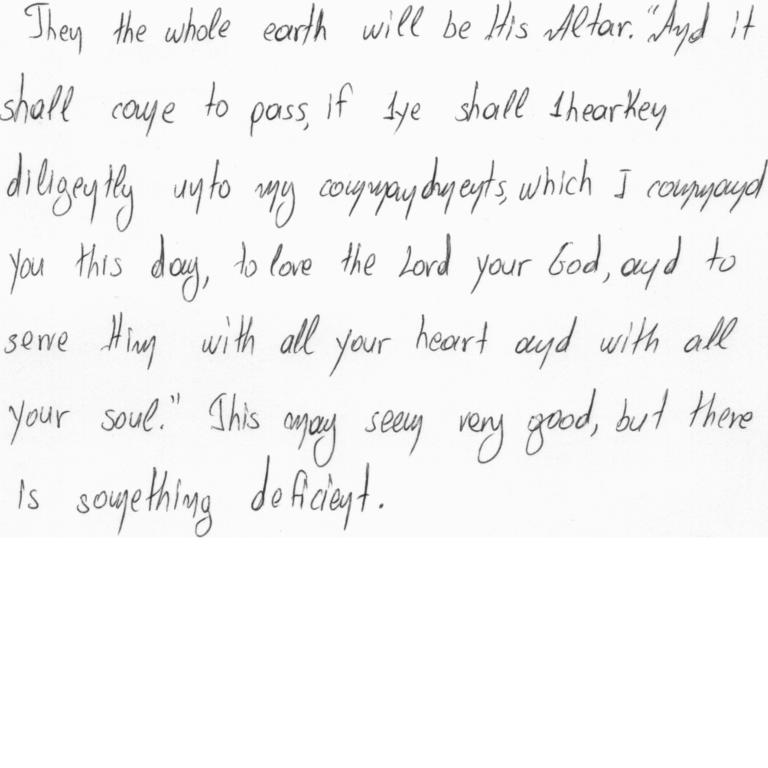}}
        \caption{Style Input}
    \end{subfigure}
    \hspace{0.1\textwidth}
    \begin{subfigure}{0.4\textwidth}
        \fbox{\includegraphics[width=\textwidth]{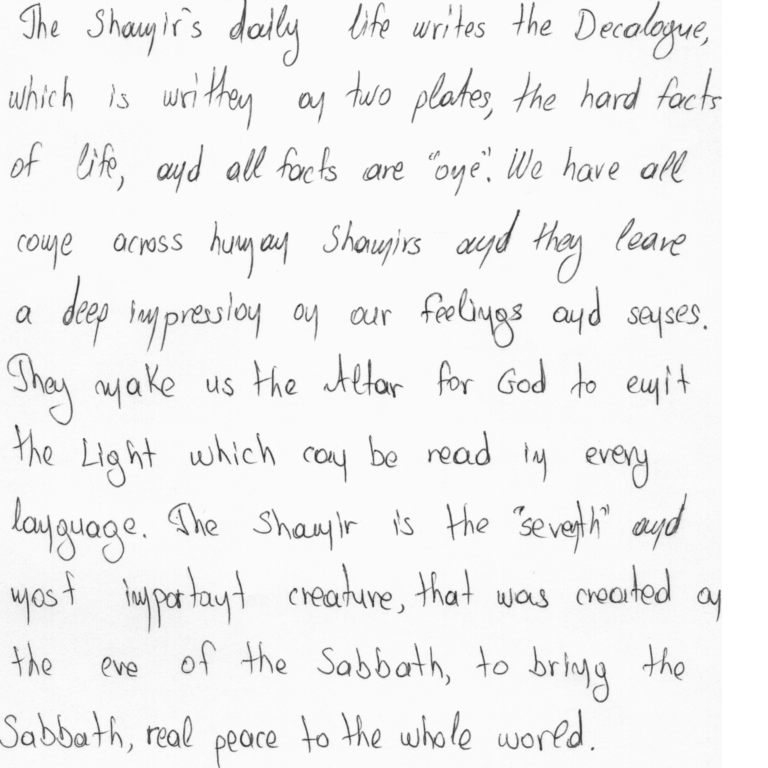}}
        \caption{Genuine}
    \end{subfigure}
    \hspace{0.1\textwidth}
    \begin{subfigure}{0.4\textwidth}
        \fbox{\includegraphics[width=\textwidth]{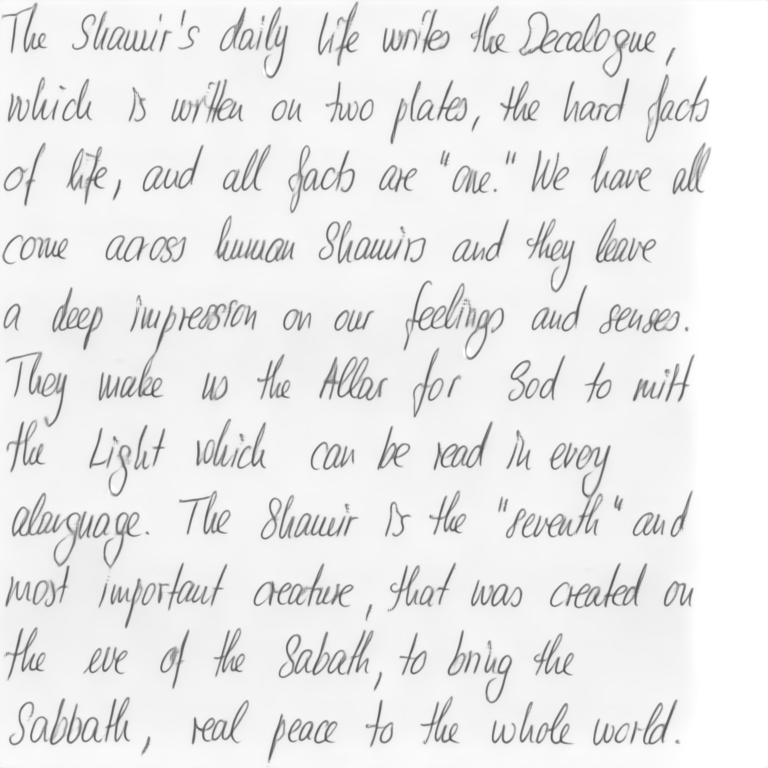}}
        \caption{Ours}
    \end{subfigure}
    \hspace{0.1\textwidth}
    \begin{subfigure}{0.4\textwidth}
        \fbox{\includegraphics[width=\textwidth]{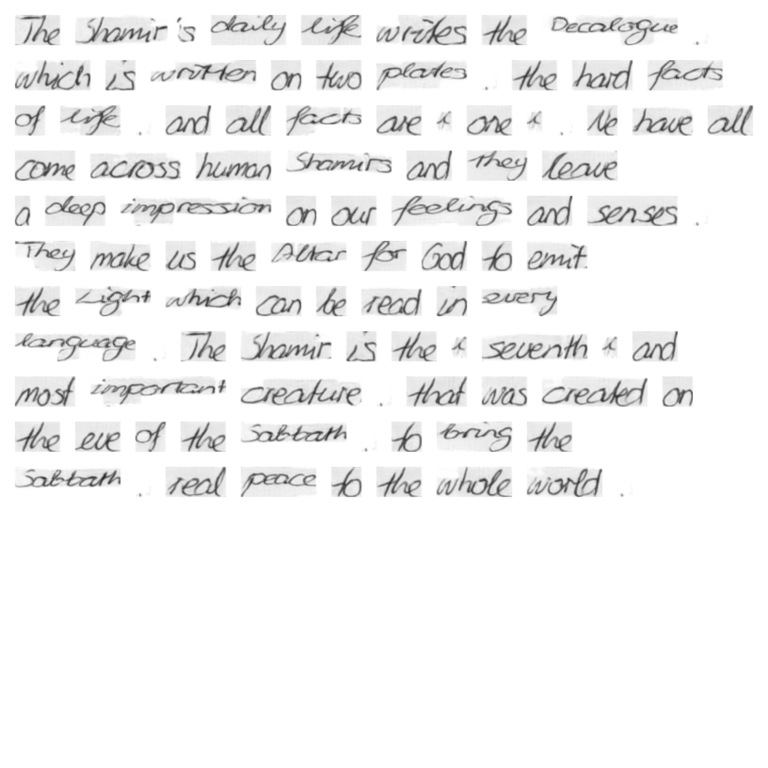}}
        \caption{VATr}
    \end{subfigure}
    \hspace{0.1\textwidth}
    \begin{subfigure}{0.4\textwidth}
        \fbox{\includegraphics[width=\textwidth]{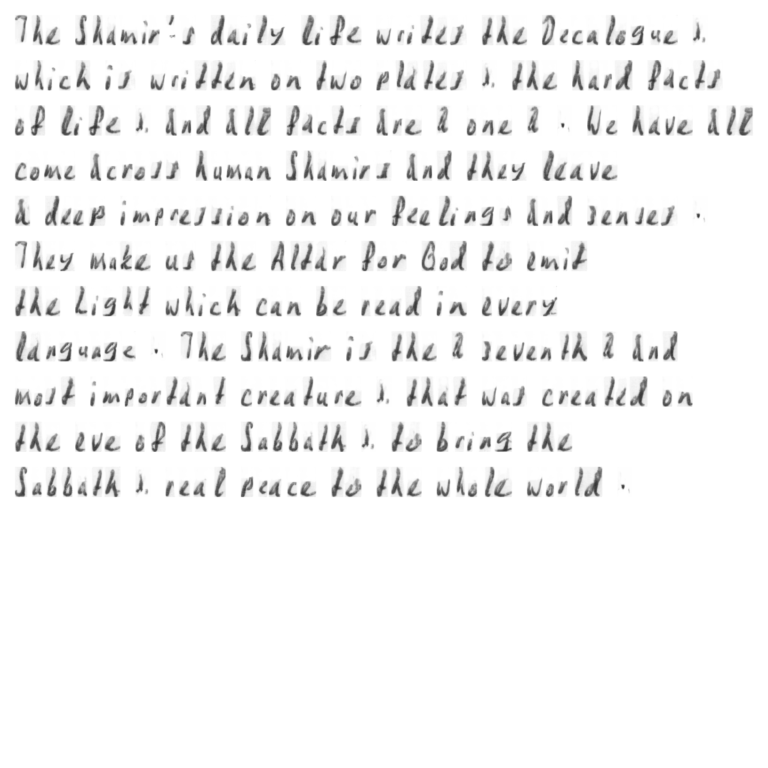}}
        \caption{HiGAN+}
    \end{subfigure}
    \hspace{0.1\textwidth}
    \begin{subfigure}{0.4\textwidth}
        \fbox{\includegraphics[width=\textwidth]{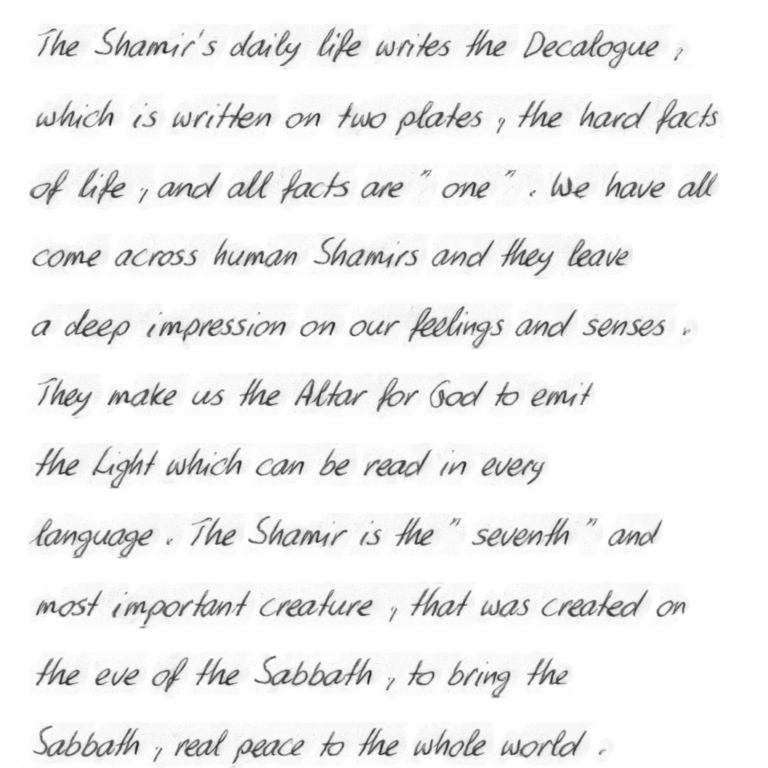}}
        \caption{TS-GAN}
    \end{subfigure}
    \caption{Image 'd04-032' from IAM. Display of the style input (a), the genuine paragraph (b), and the outputs of different imitation approaches (c-f).}
    \label{fig:sample1}
\end{figure}

\begin{figure}[hp]
    \centering
    \begin{subfigure}{0.40\textwidth}
        \fbox{\includegraphics[width=\textwidth]{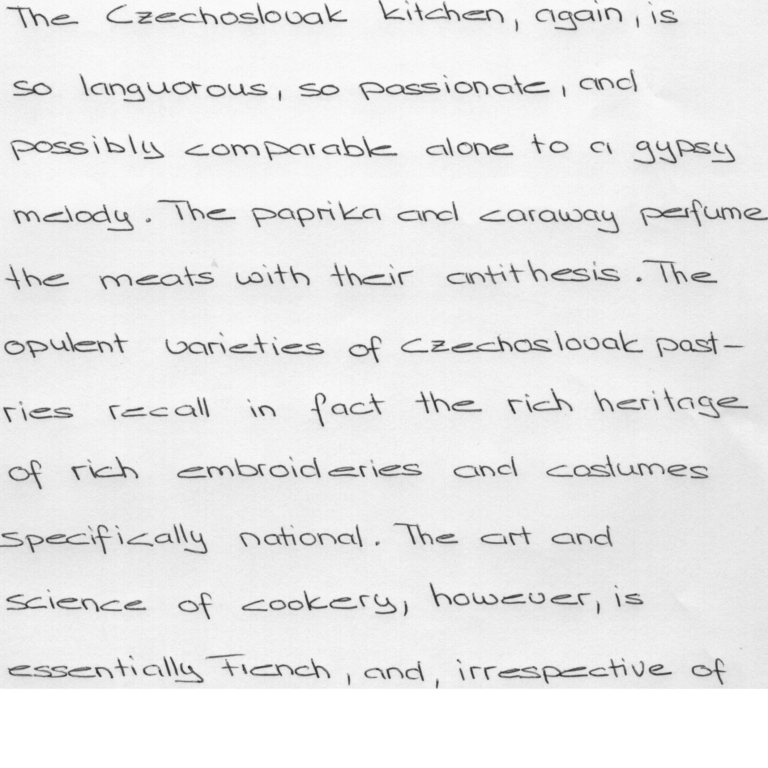}}
        \caption{Style Input}
    \end{subfigure}
    \hspace{0.1\textwidth}
    \begin{subfigure}{0.40\textwidth}
        \fbox{\includegraphics[width=\textwidth]{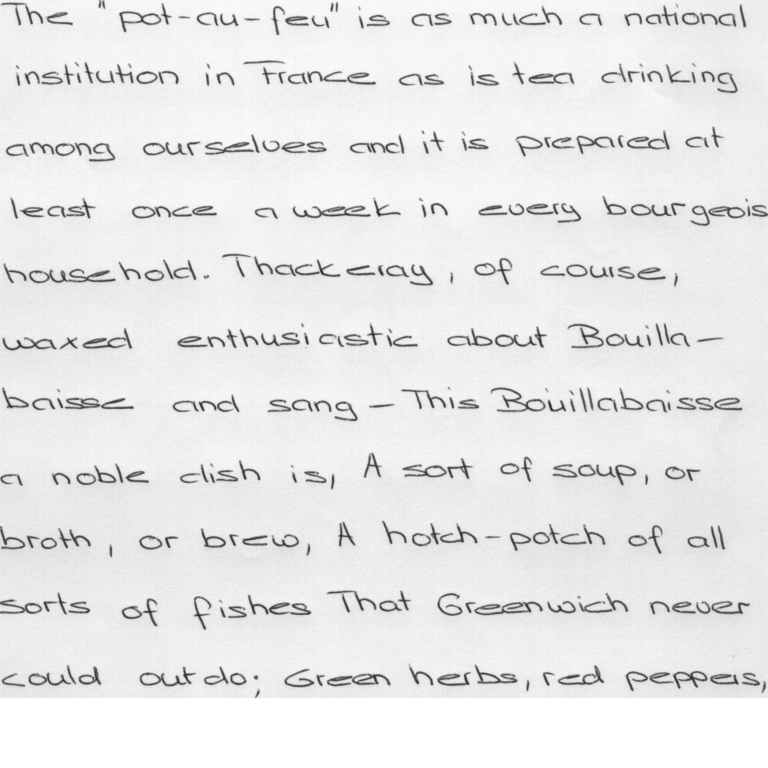}}
        \caption{Genuine}
    \end{subfigure}
    \hspace{0.1\textwidth}
    \begin{subfigure}{0.40\textwidth}
        \fbox{\includegraphics[width=\textwidth]{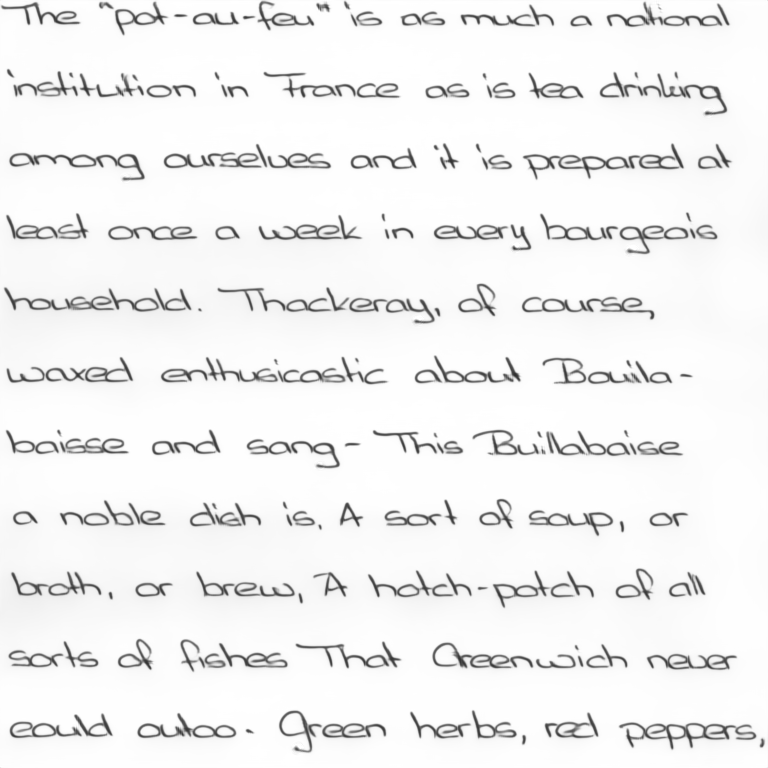}}
        \caption{Ours}
    \end{subfigure}
    \hspace{0.1\textwidth}
    \begin{subfigure}{0.40\textwidth}
        \fbox{\includegraphics[width=\textwidth]{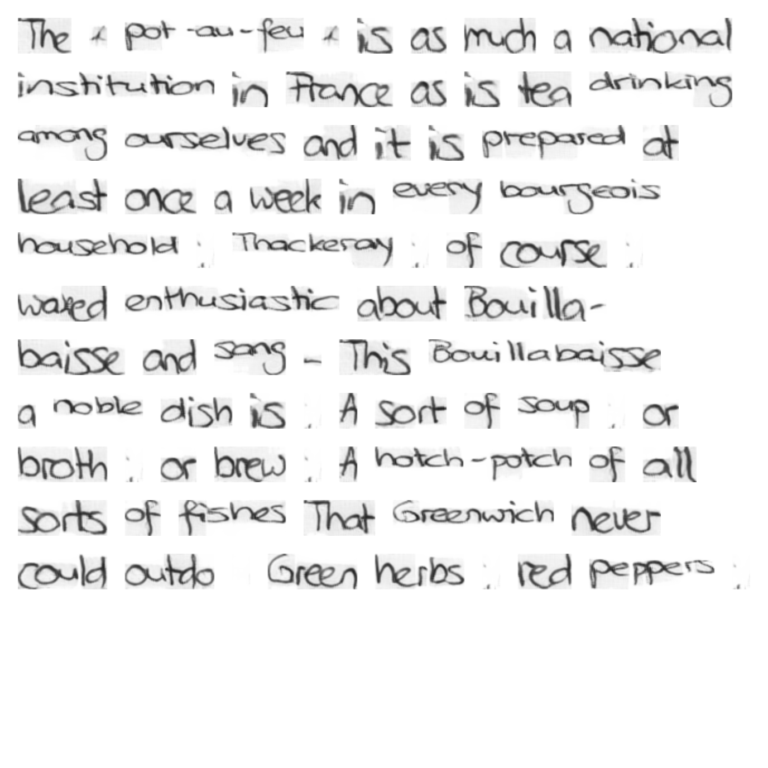}}
        \caption{VATr}
    \end{subfigure}
    \hspace{0.1\textwidth}
    \begin{subfigure}{0.40\textwidth}
        \fbox{\includegraphics[width=\textwidth]{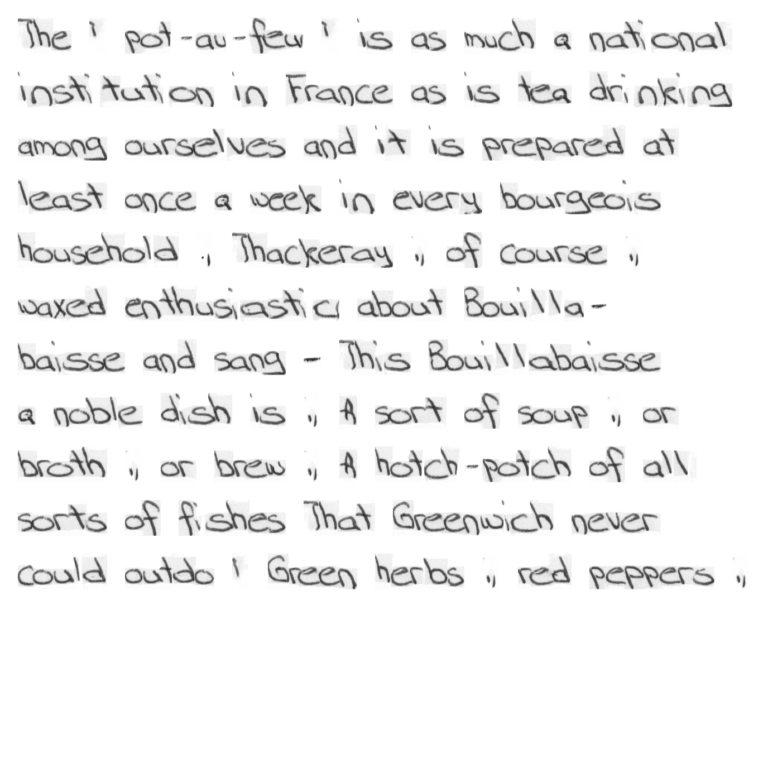}}
        \caption{HiGAN+}
    \end{subfigure}
    \hspace{0.1\textwidth}
    \begin{subfigure}{0.40\textwidth}
        \fbox{\includegraphics[width=\textwidth]{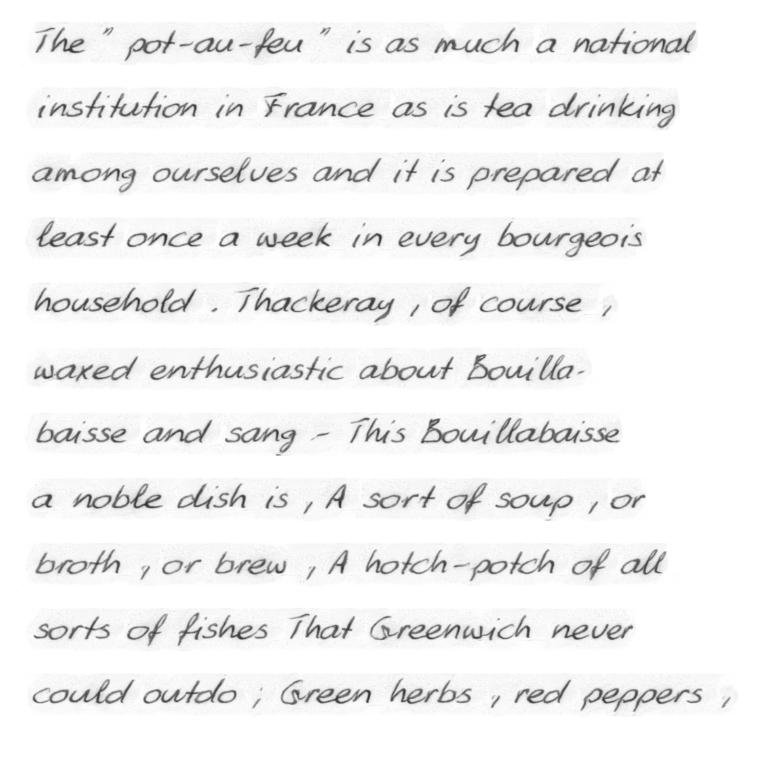}}
        \caption{TS-GAN}
    \end{subfigure}
    \caption{Image 'f07-084a' from IAM. Display of the style input (a), the genuine paragraph (b), and the outputs of different imitation approaches (c-f).}
    \label{fig:sample0}
\end{figure}

\begin{figure}[hp]
    \centering
    \begin{subfigure}{0.40\textwidth}
        \fbox{\includegraphics[width=\textwidth]{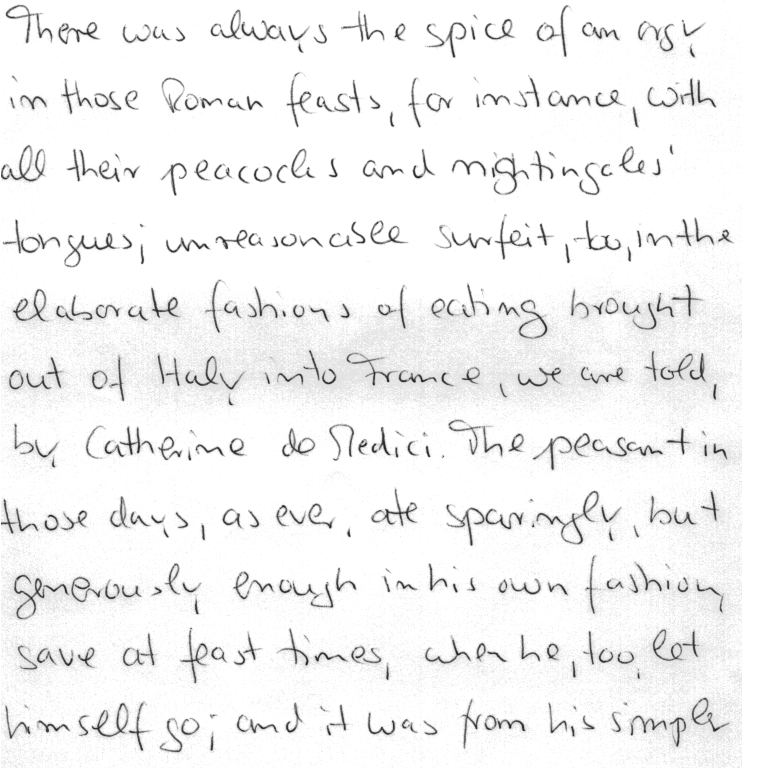}}
        \caption{Style Input}
    \end{subfigure}
    \hspace{0.1\textwidth}
    \begin{subfigure}{0.40\textwidth}
        \fbox{\includegraphics[width=\textwidth]{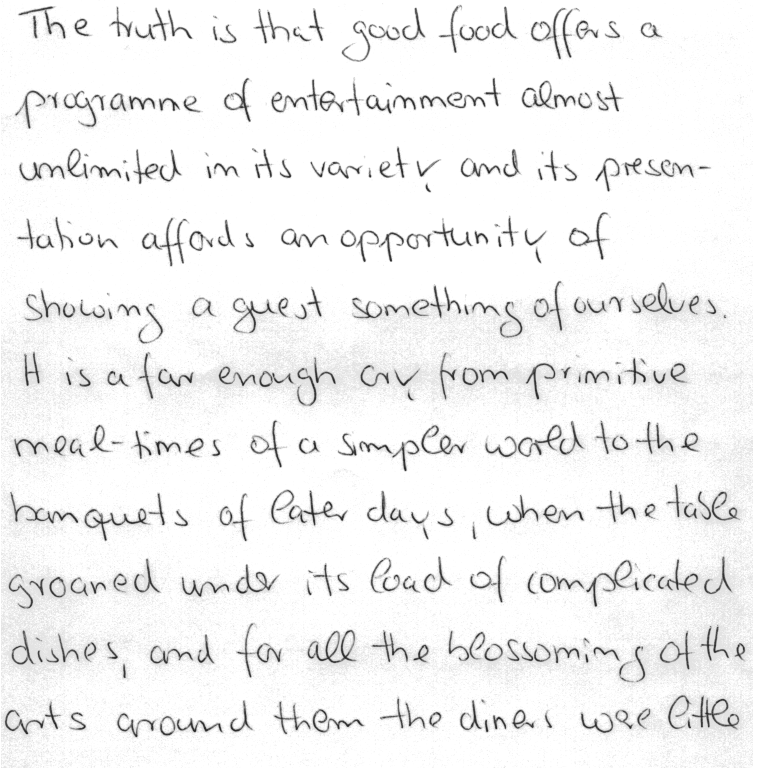}}
        \caption{Genuine}
    \end{subfigure}
    \hspace{0.1\textwidth}
    \begin{subfigure}{0.40\textwidth}
        \fbox{\includegraphics[width=\textwidth]{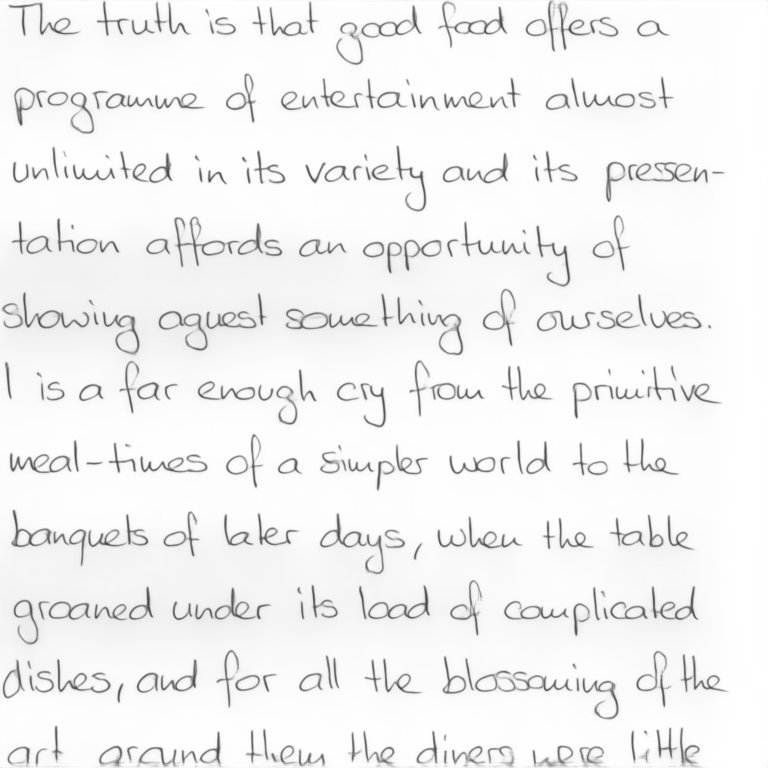}}
        \caption{Ours}
    \end{subfigure}
    \hspace{0.1\textwidth}
    \begin{subfigure}{0.40\textwidth}
        \fbox{\includegraphics[width=\textwidth]{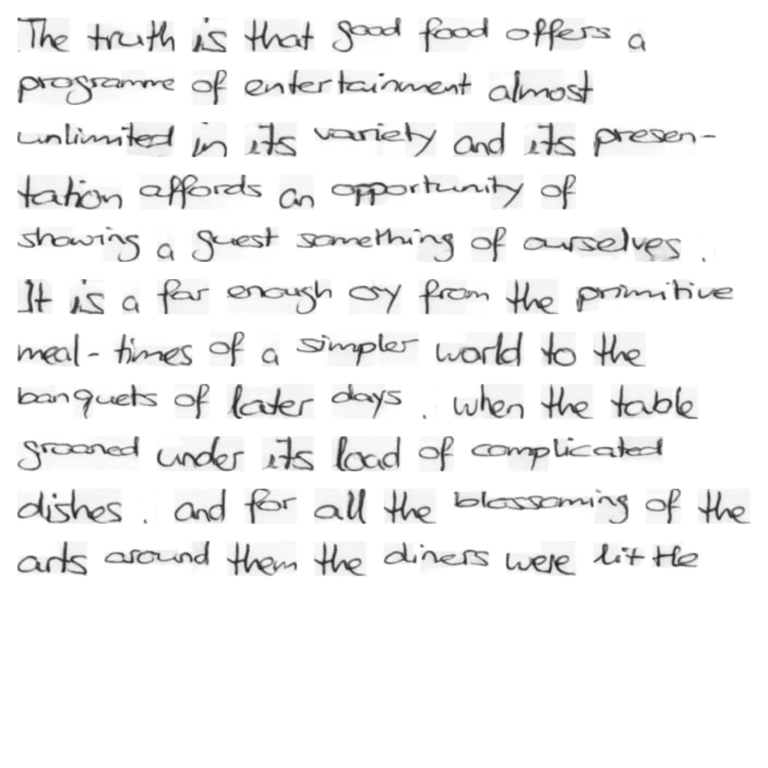}}
        \caption{VATr}
    \end{subfigure}
    \hspace{0.1\textwidth}
    \begin{subfigure}{0.40\textwidth}
        \fbox{\includegraphics[width=\textwidth]{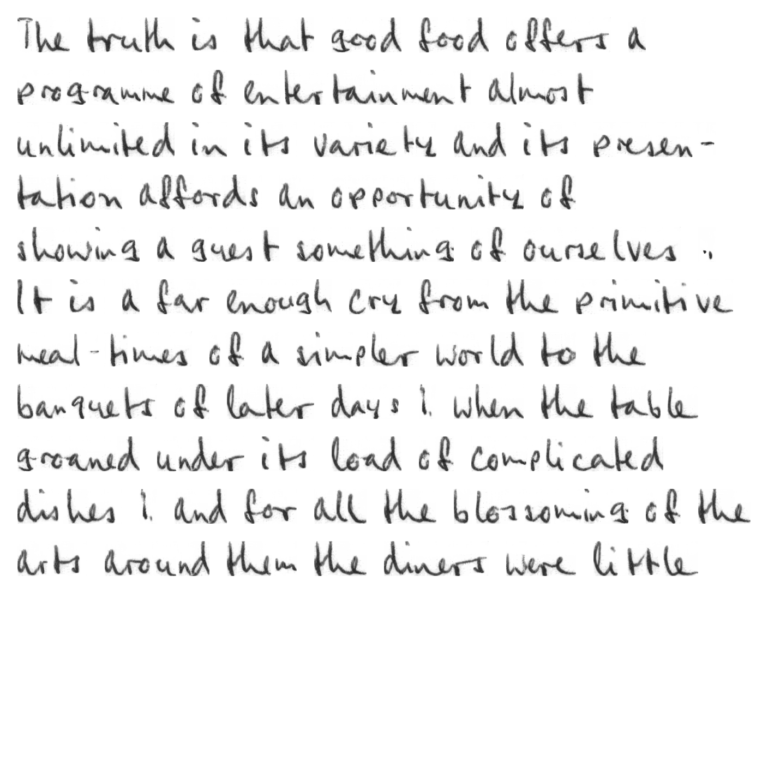}}
        \caption{HiGAN+}
    \end{subfigure}
    \hspace{0.1\textwidth}
    \begin{subfigure}{0.40\textwidth}
        \fbox{\includegraphics[width=\textwidth]{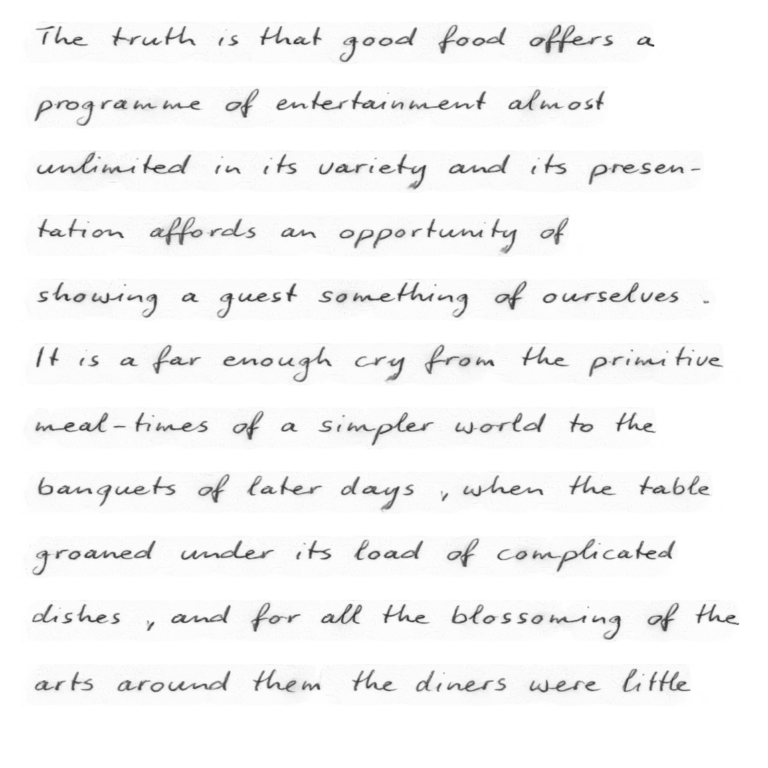}}
        \caption{TS-GAN}
    \end{subfigure}
    \caption{Image 'f07-013' from IAM. Display of the style input (a), the genuine paragraph (b), and the outputs of different imitation approaches (c-f).}
    \label{fig:sample2}
\end{figure}

\begin{figure}[hp]
    \centering
    \begin{subfigure}{0.40\textwidth}
        \fbox{\includegraphics[width=\textwidth]{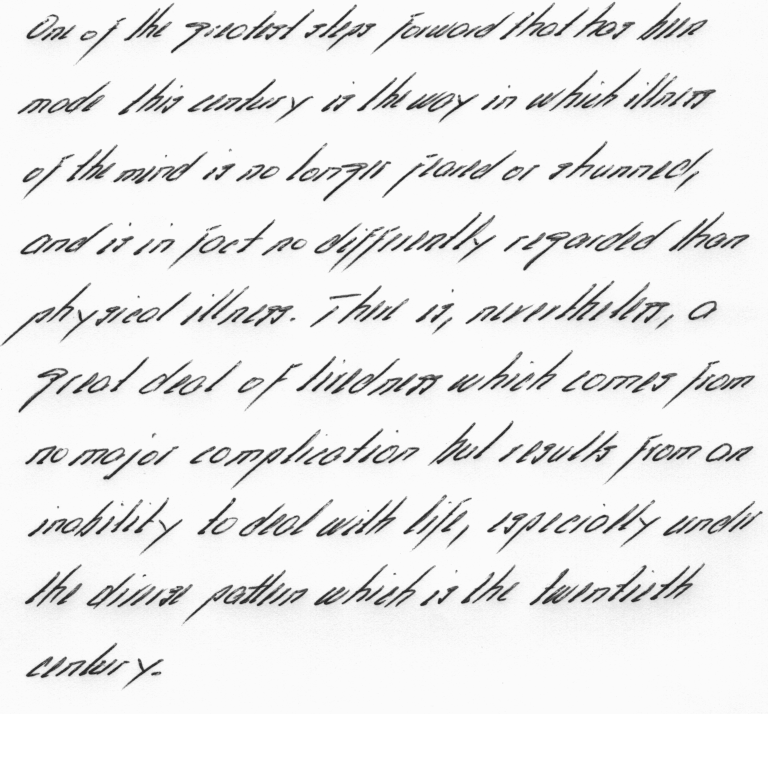}}
        \caption{Style Input}
    \end{subfigure}
    \hspace{0.1\textwidth}
    \begin{subfigure}{0.40\textwidth}
        \fbox{\includegraphics[width=\textwidth]{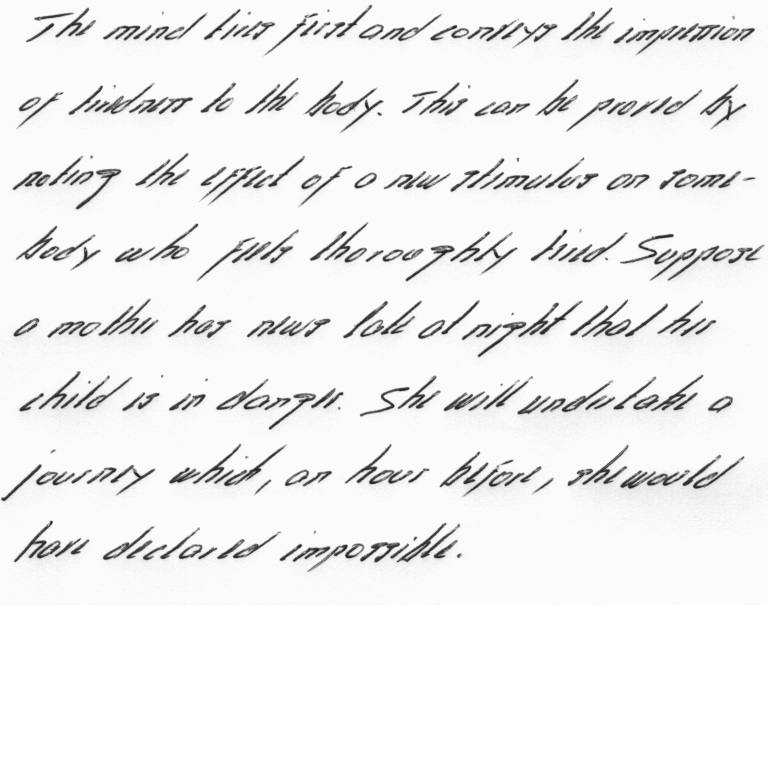}}
        \caption{Genuine}
    \end{subfigure}
    \hspace{0.1\textwidth}
    \begin{subfigure}{0.40\textwidth}
        \fbox{\includegraphics[width=\textwidth]{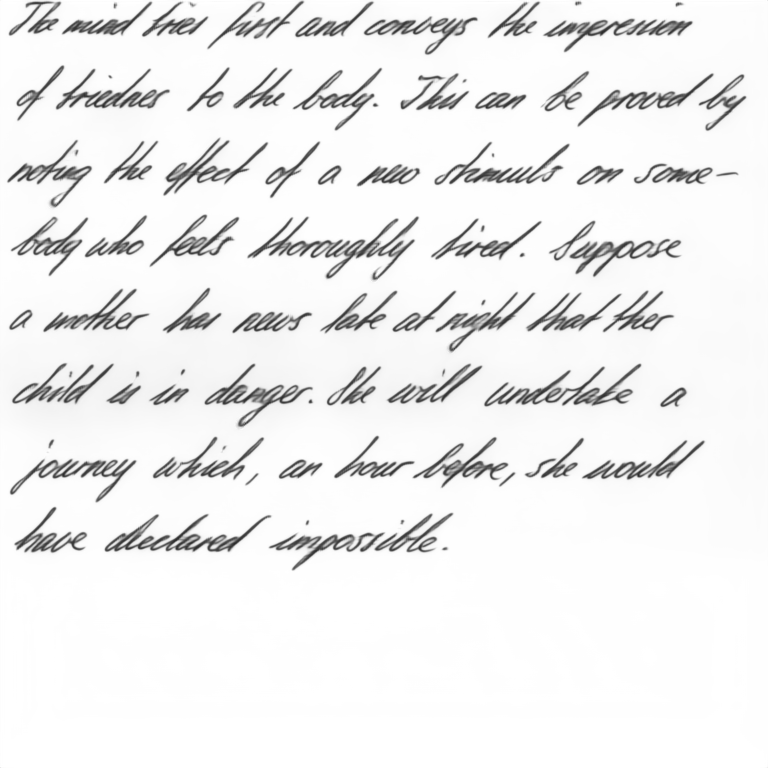}}
        \caption{Ours}
    \end{subfigure}
    \hspace{0.1\textwidth}
    \begin{subfigure}{0.40\textwidth}
        \fbox{\includegraphics[width=\textwidth]{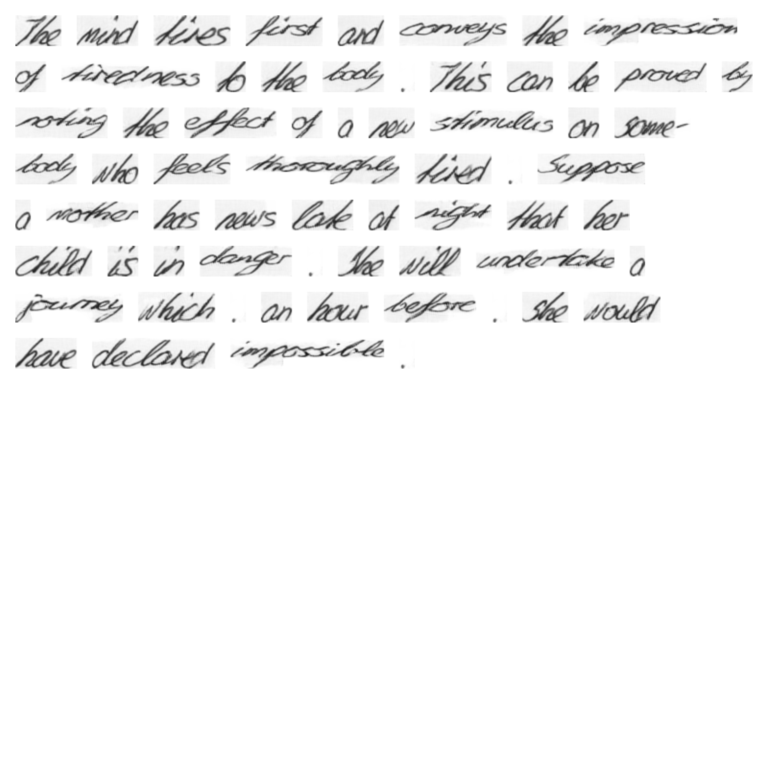}}
        \caption{VATr}
    \end{subfigure}
    \hspace{0.1\textwidth}
    \begin{subfigure}{0.40\textwidth}
        \fbox{\includegraphics[width=\textwidth]{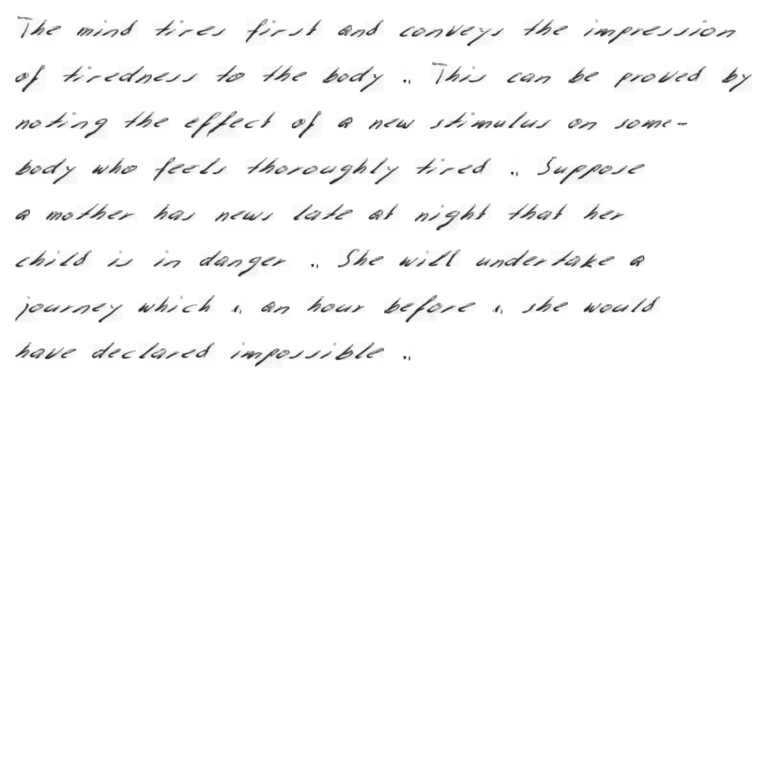}}
        \caption{HiGAN+}
    \end{subfigure}
    \hspace{0.1\textwidth}
    \begin{subfigure}{0.40\textwidth}
        \fbox{\includegraphics[width=\textwidth]{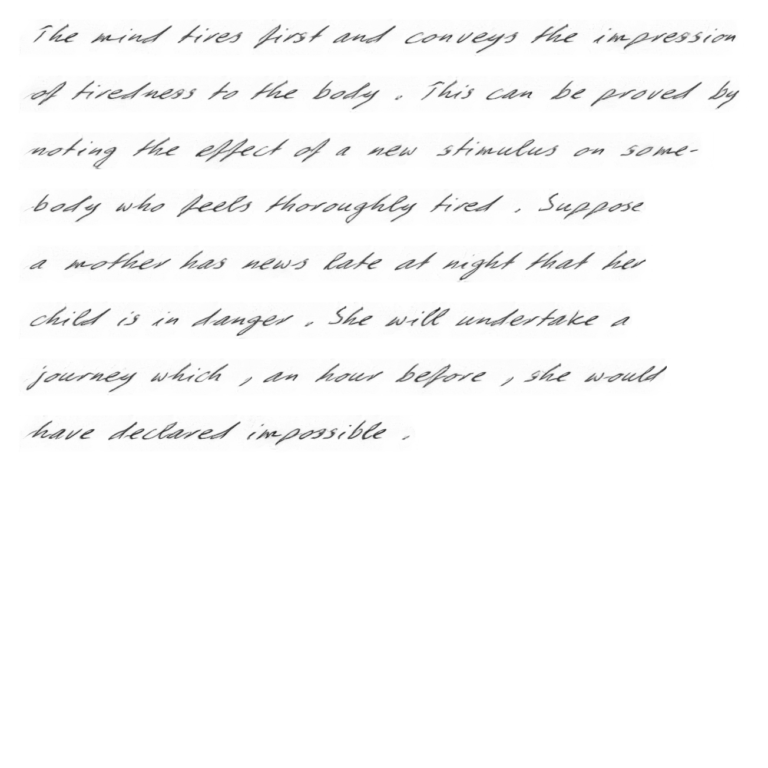}}
        \caption{TS-GAN}
    \end{subfigure}
    \caption{Image 'd06-041' from IAM. Display of the style input (a), the genuine paragraph (b), and the outputs of different imitation approaches (c-f).}
    \label{fig:sample3}
\end{figure}

\Cref{fig:sample1,fig:sample0,fig:sample2,fig:sample3} show qualitative results of the samples 'd04-032', 'f07-084a', 'f07-013', and 'd06-041' from the IAM database~\citep{marti2002iam}. We give the style input, the genuine, and the synthetically generated paragraphs. 'd04-032' has a very unique writing style. HiGAN+~\citep{gan2022higanplus} has problems replicating the style. VATr~\citep{pippi2023archetypes} and TS-GAN~\citep{davis2020text} generate better results but still not preserve the writer's style. Our approach is the closest but introduces a light background artefact.

'f07-084a' is quite an unusual sample because the slant is more leaned towards the left than towards the right. TS-GAN does not adapt to this style and is closer to the previous style than to this one. VATr also has problems with that style. HiGAN+ is the best comparison approach but is making some errors, e.g., ``,'' are often ``,,''. 
Also, the strokes of VATr and HiGAN+ are not smooth.
By contrast, our model replicates the style quite well but forgot an ``l'' in ``Bouilla-baisse''.

\begin{figure}[hp]
    \centering
    \begin{subfigure}{0.47\textwidth}
        \fbox{\includegraphics[width=\textwidth]{img_f07-084a_style_f07-076a.png}}
        \caption{Style Input}
    \end{subfigure}
    \hfill
    \begin{subfigure}{0.47\textwidth}
        \fbox{\includegraphics[width=\textwidth]{img_f07-084a_ours.png}}
        \caption{Ours}
    \end{subfigure}
    \begin{subfigure}{0.47\textwidth}
        \fbox{\includegraphics[width=\textwidth]{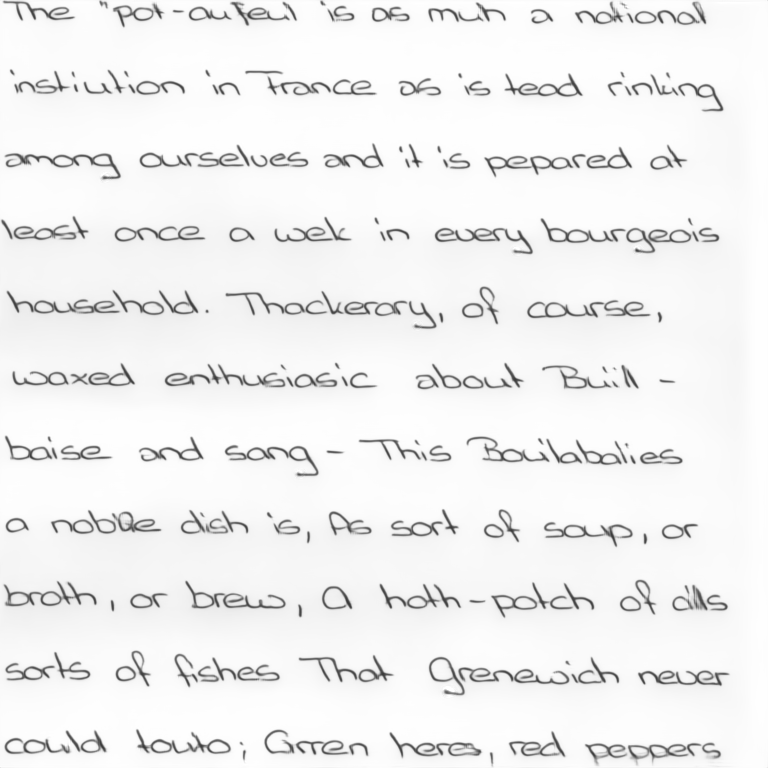}}
        \caption{Ours - Cosine}
    \end{subfigure}
    \hfill
    \begin{subfigure}{0.47\textwidth}
        \fbox{\includegraphics[width=\textwidth]{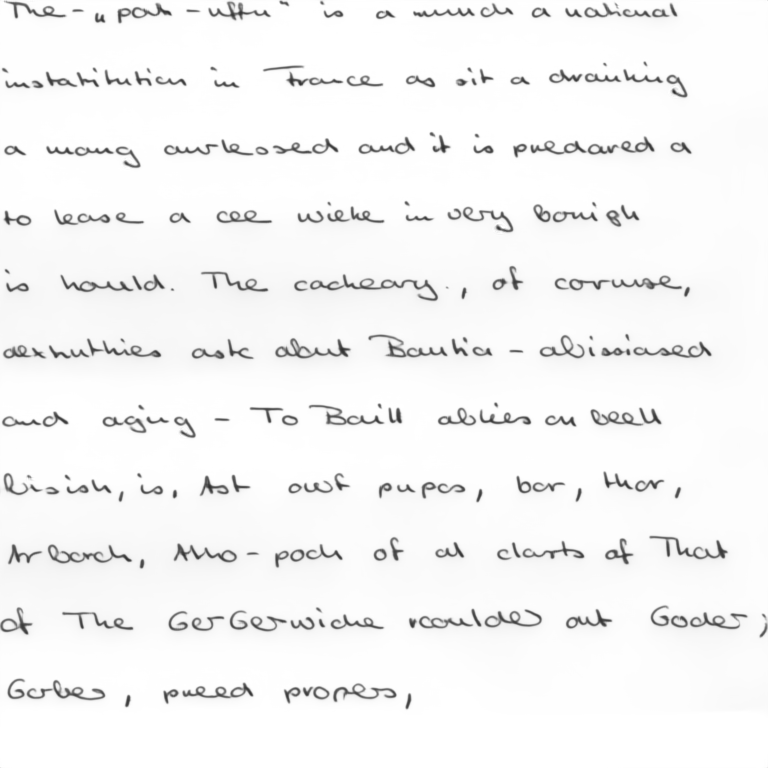}}
        \caption{Ours - no NL}
    \end{subfigure}
    \caption{Image 'f07-084a' from IAM~\citep{marti2002iam}. Display of the style input (a), and the outputs of our different imitation approaches (b-d).}
    \label{fig:sample0_ours}
\end{figure}

\begin{figure}[hp]
    \centering
    \begin{subfigure}{0.47\textwidth}
        \fbox{\includegraphics[width=\textwidth]{img_d04_032_style_d04_037.png}}
        \caption{Style Input}
    \end{subfigure}
    \hfill
    \begin{subfigure}{0.47\textwidth}
        \fbox{\includegraphics[width=\textwidth]{img_d04_032_ours.png}}
        \caption{Ours}
    \end{subfigure}
    \begin{subfigure}{0.47\textwidth}
        \fbox{\includegraphics[width=\textwidth]{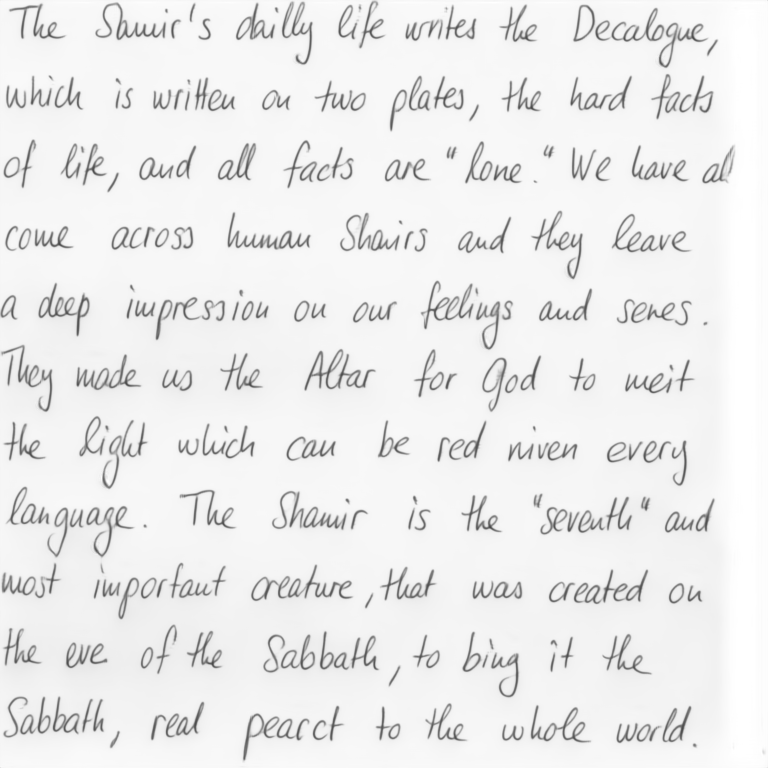}}
        \caption{Ours - Cosine}
    \end{subfigure}
    \hfill
    \begin{subfigure}{0.47\textwidth}
        \fbox{\includegraphics[width=\textwidth]{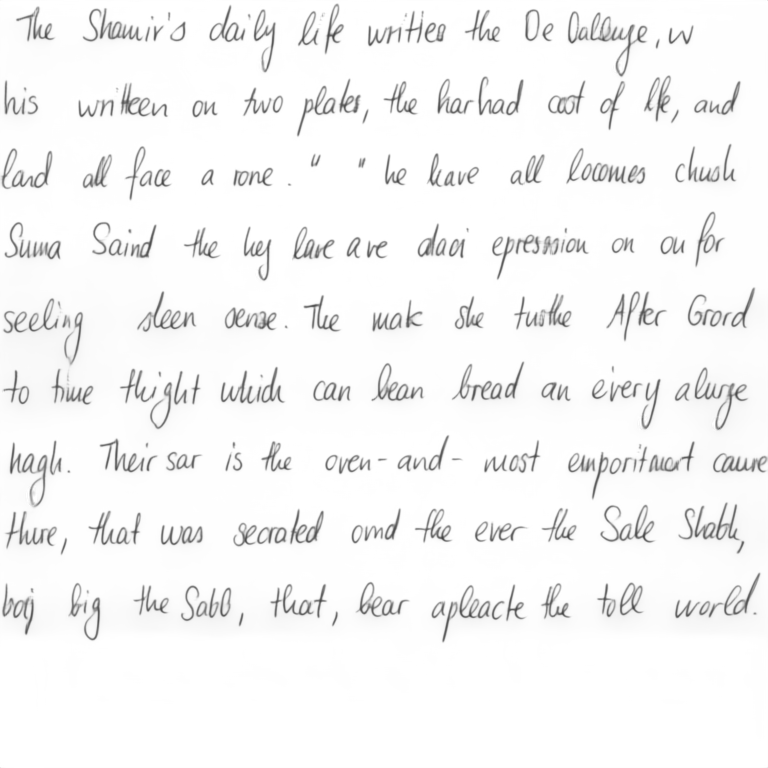}}
        \caption{Ours - no NL}
    \end{subfigure}
    \caption{Image 'd04-032' from IAM~\citep{marti2002iam}. Display of the style input (a), and the outputs of our different imitation approaches (b-d).}
    \label{fig:sample1_ours}
\end{figure}

Our three different diffusion variants in \cref{fig:sample0_ours} and \cref{fig:sample1_ours} reflect the trends shown in the main paper. The style is quite similarly preserved for all the variants, but the content often differs from the target text for ``Ours - Cosine'' and especially for ``Ours - no NL''.

\end{appendices}

\clearpage

\bibliography{sn-bibliography}%

@String(PAMI = {IEEE Trans. Pattern Anal. Mach. Intell.})

@String(CVPR= {IEEE Conf. Comput. Vis. Pattern Recog.})

@String(ICCV= {Int. Conf. Comput. Vis.})

@String(ECCV= {Eur. Conf. Comput. Vis.})

@String(BMVC= {Brit. Mach. Vis. Conf.})

@String(ICLR = {Int. Conf. Learn. Represent.})

@String(PAMI  = {IEEE TPAMI})

@String(CVPR  = {CVPR})

@String(ICCV  = {ICCV})

@String(ECCV  = {ECCV})

@String(BMVC  =	{BMVC})

@String(ICLR  = {ICLR})

@inproceedings{mayr2020spatio,
	title        = {Spatio-Temporal Handwriting Imitation},
	author       = {Mayr, Martin and Stumpf, Martin and Nicolaou, Anguelos and Seuret, Mathias and Maier, Andreas and Christlein, Vincent},
	year         = 2020,
	booktitle    = {European Conference on Computer Vision (ECCV) Workshops},
	publisher    = {Springer International Publishing},
	address      = {Cham},
	pages        = {528--543},
	isbn         = {978-3-030-68238-5},
	editor       = {Bartoli, Adrien and Fusiello, Andrea},
    doi = {10.1007/978-3-030-68238-5_38}
}

@inproceedings{kang2020ganwriting,
	title        = {GANwriting: Content-Conditioned Generation of Styled Handwritten Word Images},
	author       = {Kang, Lei and Riba, Pau and Wang, Yaxing and Rusi{\~{n}}ol, Mar{\c{c}}al and Forn{\'e}s, Alicia and Villegas, Mauricio},
	year         = 2020,
	booktitle    = {European Conference on Computer Vision (ECCV)},
	publisher    = {Springer International Publishing},
	address      = {Cham},
	pages        = {273--289},
	isbn         = {978-3-030-58592-1},
	editor       = {Vedaldi, Andrea and Bischof, Horst and Brox, Thomas and Frahm, Jan-Michael},
    doi = {10.1007/978-3-030-58592-1_17}
}

@inproceedings{davis2020text,
	title        = {Text and Style Conditioned {GAN} for Generation of Offline Handwriting Lines},
	author       = {Brian Davis and Chris Tensmeyer and Brian Price and Curtis Wigington and Bryan Morse and Rajiv Jain},
	year         = 2020,
	booktitle    = {British Machine Vision Conference (BMVC)},
	url          = {https://www.bmvc2020-conference.com/assets/papers/0815.pdf},
	eprint       = {2009.00678},
	archiveprefix = {arXiv},
	primaryclass = {cs.CV}
}

@inproceedings{dhariwal2021guidance,
	title        = {Diffusion Models Beat {GANs} on Image Synthesis},
	author       = {Dhariwal, Prafulla and Nichol, Alexander},
	year         = 2021,
	booktitle    = {Advances in Neural Information Processing Systems (NeurIPS)},
	publisher    = {Curran Associates, Inc.},
	volume       = 34,
	pages        = {8780--8794},
	editor       = {M. Ranzato and A. Beygelzimer and Y. Dauphin and P.S. Liang and J. Wortman Vaughan}
}

@inproceedings{ho2021classifierfree,
	title        = {Classifier-Free Diffusion Guidance},
	author       = {Jonathan Ho and Tim Salimans},
	year         = 2021,
	booktitle    = {Advances in Neural Information Processing Systems (NeurIPS) Workshop on Deep Generative Models and Downstream Applications},
	url          = {https://openreview.net/forum?id=qw8AKxfYbI}
}

@inproceedings{ho2020dm,
	title        = {Denoising Diffusion Probabilistic Models},
	author       = {Ho, Jonathan and Jain, Ajay and Abbeel, Pieter},
	year         = 2020,
	booktitle    = {Advances in Neural Information Processing Systems (NeurIPS)},
	publisher    = {Curran Associates, Inc.},
	volume       = 33,
	pages        = {6840--6851},
	editor       = {H. Larochelle and M. Ranzato and R. Hadsell and M.F. Balcan and H. Lin}
}

@article{kang2021content,
	title        = {Content and Style Aware Generation of Text-line Images for Handwriting Recognition},
	author       = {Kang, Lei and Riba, Pau and Rusinol, Marcal and Fornes, Alicia and Villegas, Mauricio},
	year         = 2021,
	journal      = {IEEE PAMI},
	volume       = {},
	number       = {},
	pages        = {1--1},
	doi          = {10.1109/TPAMI.2021.3122572 }
}

@article{kang2022pay,
	title        = {Pay attention to what you read: Non-recurrent handwritten text-Line recognition},
	author       = {Lei Kang and Pau Riba and Marçal Rusiñol and Alicia Fornés and Mauricio Villegas},
	year         = 2022,
	journal      = {Pattern Recognit.},
	volume       = 129,
	pages        = 108766,
	doi          = {10.1016/j.patcog.2022.108766},
	issn         = {0031-3203},
}

@InProceedings{wick2021bidirectional,
author="Wick, Christoph
and Z{\"o}llner, Jochen
and Gr{\"u}ning, Tobias",
editor="Llad{\'o}s, Josep
and Lopresti, Daniel
and Uchida, Seiichi",
title="Transformer for Handwritten Text Recognition Using Bidirectional Post-decoding",
booktitle="Document Analysis and Recognition -- ICDAR 2021",
year="2021",
publisher="Springer International Publishing",
address="Cham",
pages="112--126",
doi="https://doi.org/10.1007/978-3-030-86334-0_8",
isbn="978-3-030-86334-0"
}

@inproceedings{mattick2021smartpatch,
	title        = {SmartPatch: Improving Handwritten Word Imitation with Patch Discriminators},
	author       = {Mattick, Alexander and Mayr, Martin and Seuret, Mathias and Maier, Andreas and Christlein, Vincent},
	year         = 2021,
	booktitle    = {Document Analysis and Recognition (ICDAR)},
	publisher    = {Springer International Publishing},
	address      = {Cham},
	pages        = {268--283},
	isbn         = {978-3-030-86549-8},
	editor       = {Llad{\'o}s, Josep and Lopresti, Daniel and Uchida, Seiichi},
    doi = {10.1007/978-3-030-86549-8_18}
}

@inproceedings{fogel2020scrabble,
	title        = {ScrabbleGAN: Semi-Supervised Varying Length Handwritten Text Generation},
	author       = {Fogel, Sharon and Averbuch-Elor, Hadar and Cohen, Sarel and Mazor, Shai and Litman, Roee},
	year         = 2020,
	month        = {Jun},
	booktitle    = {IEEE/CVF Conference on Computer Vision and Pattern Recognition (CVPR)},
    pages={4323-4332},
    doi = {10.1109/CVPR42600.2020.00438}
}

@inproceedings{ramesh2021dalle,
	title        = {Zero-Shot Text-to-Image Generation},
	author       = {Ramesh, Aditya and Pavlov, Mikhail and Goh, Gabriel and Gray, Scott and Voss, Chelsea and Radford, Alec and Chen, Mark and Sutskever, Ilya},
	year         = 2021,
	month        = {Jul},
	booktitle    = {International Conference on Machine Learning (ICML)},
	publisher    = {PMLR},
	volume       = 139,
	pages        = {8821--8831},
	url          = {https://proceedings.mlr.press/v139/ramesh21a.html},
	editor       = {Meila, Marina and Zhang, Tong},
}

@inproceedings{dosovitskiy2021vit,
	title        = {An Image is Worth 16x16 Words: Transformers for Image Recognition at Scale},
	author       = {Alexey Dosovitskiy and Lucas Beyer and Alexander Kolesnikov and Dirk Weissenborn and Xiaohua Zhai and Thomas Unterthiner and Mostafa Dehghani and Matthias Minderer and Georg Heigold and Sylvain Gelly and Jakob Uszkoreit and Neil Houlsby},
	year         = 2021,
	booktitle    = {International Conference on Learning Representations (ICLR)},
	url          = {https://openreview.net/forum?id=YicbFdNTTy}
}

@misc{graves2014generating,
      title={Generating Sequences With Recurrent Neural Networks}, 
      author={Alex Graves},
      year={2014},
      eprint={1308.0850},
      archivePrefix={arXiv},
      primaryClass={cs.NE},
      url={https://arxiv.org/abs/1308.0850}, 
}

@article{yang2023dmsurvey,
	title        = {Diffusion Models: A Comprehensive Survey of Methods and Applications},
	author       = {Yang, Ling and Zhang, Zhilong and Song, Yang and Hong, Shenda and Xu, Runsheng and Zhao, Yue and Zhang, Wentao and Cui, Bin and Yang, Ming-Hsuan},
	year         = 2023,
	month        = {Nov},
	journal      = {ACM Comput. Surv.},
	publisher    = {Association for Computing Machinery},
	address      = {New York, NY, USA},
	volume       = 56,
	number       = 4,
	doi          = {10.1145/3626235},
	issn         = {0360-0300},
	issue_date   = {April 2024},
	articleno    = 105,
	numpages     = 39,
}

@inproceedings{sohl2015dm,
	title        = {Deep Unsupervised Learning using Nonequilibrium Thermodynamics},
	author       = {Sohl-Dickstein, Jascha and Weiss, Eric and Maheswaranathan, Niru and Ganguli, Surya},
	year         = 2015,
	month        = {Jul},
	booktitle    = {International Conference on Machine Learning (ICML)},
	publisher    = {PMLR},
	address      = {Lille, France},
	volume       = 37,
	pages        = {2256--2265},
	url          = {https://proceedings.mlr.press/v37/sohl-dickstein15.html},
	editor       = {Bach, Francis and Blei, David},
}

@inproceedings{song2021denoising,
	title        = {Denoising Diffusion Implicit Models},
	author       = {Jiaming Song and Chenlin Meng and Stefano Ermon},
	year         = 2021,
	booktitle    = {International Conference on Learning Representations (ICLR)},
	url          = {https://openreview.net/forum?id=St1giarCHLP}
}

@inproceedings{kingma2014vae,
	title        = {Auto-Encoding Variational Bayes},
	author       = {Kingma, Diederik P. and Welling, Max},
	year         = 2014,
	booktitle    = {International Conference on Learning Representations (ICLR)}
}

@article{bisio2016kinematics,
	title        = {Evaluation of Handwriting Movement Kinematics: From an Ecological to a Magnetic Resonance Environment},
	author       = {Bisio, Ambra and Pedullà, Ludovico and Bonzano, Laura and Ruggeri, Piero and Brichetto, Giampaolo and Bove, Marco},
	year         = 2016,
	journal      = {Front. Hum. Neurosci.},
	volume       = 10,
	doi          = {10.3389/fnhum.2016.00488},
	issn         = {1662-5161},
}

@inproceedings{rezende2014vae,
	title        = {Stochastic Backpropagation and Approximate Inference in Deep Generative Models},
	author       = {Rezende, Danilo Jimenez and Mohamed, Shakir and Wierstra, Daan},
	year         = 2014,
	month        = {Jun},
	booktitle    = {International Conference on Machine Learning (ICML)},
	publisher    = {PMLR},
	address      = {Bejing, China},
	volume       = 32,
	number       = 2,
	pages        = {1278--1286},
	url          = {https://proceedings.mlr.press/v32/rezende14.html},
	editor       = {Xing, Eric P. and Jebara, Tony},
}

@inproceedings{zdenek2021jokergan,
	title        = {JokerGAN: Memory-Efficient Model for Handwritten Text Generation with Text Line Awareness},
	author       = {Zdenek, Jan and Nakayama, Hideki},
	year         = 2021,
	booktitle    = {ACM International Conference on Multimedia},
	location     = {Virtual Event, China},
	publisher    = {Association for Computing Machinery},
	address      = {New York, NY, USA},
	series       = {MM '21},
	pages        = {5655–5663},
	doi          = {10.1145/3474085.3475713},
	isbn         = 9781450386517,
	numpages     = 9,
}

@inproceedings{zdenek2023charactergan,
	title        = {Handwritten Text Generation with Character-Specific Encoding for Style Imitation},
	author       = {Zdenek, Jan and Nakayama, Hideki},
	year         = 2023,
	booktitle    = {Document Analysis and Recognition (ICDAR)},
	publisher    = {Springer Nature Switzerland},
	address      = {Cham},
	pages        = {313--329},
	isbn         = {978-3-031-41679-8},
	editor       = {Fink, Gernot A. and Jain, Rajiv and Kise, Koichi and Zanibbi, Richard},
    doi = {10.1007/978-3-031-41679-8_18}
}

@inproceedings{aksan2018deepwriting,
	title        = {DeepWriting: Making Digital Ink Editable via Deep Generative Modeling},
	author       = {Aksan, Emre and Pece, Fabrizio and Hilliges, Otmar},
	year         = 2018,
	booktitle    = {Conference on Human Factors in Computing Systems},
	location     = {<conf-loc>, <city>Montreal QC</city>, <country>Canada</country>, </conf-loc>},
	publisher    = {Association for Computing Machinery},
	address      = {New York, NY, USA},
	pages        = {1–14},
	doi          = {10.1145/3173574.3173779},
	isbn         = 9781450356206,
}

@misc{mcinnes2020umap,
	title        = {UMAP: Uniform Manifold Approximation and Projection for Dimension Reduction},
	author       = {Leland McInnes and John Healy and James Melville},
	year         = 2020,
	eprint       = {1802.03426},
	archiveprefix = {arXiv},
	primaryclass = {stat.ML}
}

@inproceedings{heusel2017fid,
	title        = {GANs Trained by a Two Time-Scale Update Rule Converge to a Local {Nash} Equilibrium},
	author       = {Heusel, Martin and Ramsauer, Hubert and Unterthiner, Thomas and Nessler, Bernhard and Hochreiter, Sepp},
	year         = 2017,
	booktitle    = {Advances in Neural Information Processing Systems (NeurIPS)},
	publisher    = {Curran Associates, Inc.},
	volume       = 30,
	pages        = {},
	editor       = {I. Guyon and U. Von Luxburg and S. Bengio and H. Wallach and R. Fergus and S. Vishwanathan and R. Garnett}
}

@inproceedings{binkowski2018kid,
	title        = {Demystifying {MMD} {GANs}},
	author       = {Mikołaj Bińkowski and Dougal J. Sutherland and Michael Arbel and Arthur Gretton},
	year         = 2018,
	booktitle    = {International Conference on Learning Representations (ICLR)},
	url          = {https://openreview.net/forum?id=r1lUOzWCW}
}

@inproceedings{nichol2021cosine,
	title        = {Improved Denoising Diffusion Probabilistic Models},
	author       = {Nichol, Alexander Quinn and Dhariwal, Prafulla},
	year         = 2021,
	month        = {Jul},
	booktitle    = {International Conference on Machine Learning (ICML)},
	publisher    = {PMLR},
	volume       = 139,
	pages        = {8162--8171},
	url          = {https://proceedings.mlr.press/v139/nichol21a.html},
	editor       = {Meila, Marina and Zhang, Tong},
}

@inproceedings{bhunia2021handwriting,
	title        = {Handwriting Transformers},
	author       = {Bhunia, Ankan Kumar and Khan, Salman and Cholakkal, Hisham and Anwer, Rao Muhammad and Khan, Fahad Shahbaz and Shah, Mubarak},
	year         = 2021,
	month        = {Oct},
	booktitle    = {IEEE/CVF International Conference on Computer Vision (ICCV)},
	pages        = {1086--1094},
    doi = {10.1109/ICCV48922.2021.00112}
}

@inproceedings{vaswani2017transformer,
 author = {Vaswani, Ashish and Shazeer, Noam and Parmar, Niki and Uszkoreit, Jakob and Jones, Llion and Gomez, Aidan N and Kaiser, \L ukasz and Polosukhin, Illia},
 booktitle = {Advances in Neural Information Processing Systems},
 editor = {I. Guyon and U. Von Luxburg and S. Bengio and H. Wallach and R. Fergus and S. Vishwanathan and R. Garnett},
 pages = {},
 publisher = {Curran Associates, Inc.},
 title = {Attention is All you Need},
 volume = {30},
 year = {2017}
}

@inproceedings{chang2022controllable,
	title        = {Style Equalization: Unsupervised Learning of Controllable Generative Sequence Models},
	author       = {Chang, Jen-Hao Rick and Shrivastava, Ashish and Koppula, Hema and Zhang, Xiaoshuai and Tuzel, Oncel},
	year         = 2022,
	month        = {Jul},
	booktitle    = {International Conference on Machine Learning (ICML)},
	publisher    = {PMLR},
	volume       = 162,
	pages        = {2917--2937},
	url          = {https://proceedings.mlr.press/v162/chang22a.html},
	editor       = {Chaudhuri, Kamalika and Jegelka, Stefanie and Song, Le and Szepesvari, Csaba and Niu, Gang and Sabato, Sivan},
}

@conference{rombach2022ldm,
	title        = {High-Resolution Image Synthesis with Latent Diffusion Models},
	author       = {Robin Rombach and Andreas Blattmann and Dominik Lorenz and Patrick Esser and Björn Ommer},
	year         = 2022,
	booktitle    = {IEEE/CVF Conference on Computer Vision and Pattern Recognition (CVPR)},
  pages={10674-10685},
	doi = {10.1109/CVPR52688.2022.01042}
}

@inproceedings{nikolaidou2023wordstylist,
	title        = {WordStylist: Styled Verbatim Handwritten Text Generation with Latent Diffusion Models},
	author       = {Nikolaidou, Konstantina and Retsinas, George and Christlein, Vincent and Seuret, Mathias and Sfikas, Giorgos and Smith, Elisa Barney and Mokayed, Hamam and Liwicki, Marcus},
	year         = 2023,
	booktitle    = {Document Analysis and Recognition (ICDAR)},
	publisher    = {Springer Nature Switzerland},
	address      = {Cham},
	pages        = {384--401},
	isbn         = {978-3-031-41679-8},
	editor       = {Fink, Gernot A. and Jain, Rajiv and Kise, Koichi and Zanibbi, Richard},
    doi = {10.1007/978-3-031-41679-8_22}
}

@inproceedings{pippi2023archetypes,
	title        = {Handwritten Text Generation From Visual Archetypes},
	author       = {Pippi, Vittorio and Cascianelli, Silvia and Cucchiara, Rita},
	year         = 2023,
	month        = {Jun},
	booktitle    = {IEEE/CVF Conference on Computer Vision and Pattern Recognition (CVPR)},
	pages        = {22458--22467},
    doi = {10.1109/CVPR52729.2023.02151}
}

@inproceedings{tang2022fewshot,
	title        = {Few-Shot Font Generation by Learning Fine-Grained Local Styles},
	author       = {Tang, Licheng and Cai, Yiyang and Liu, Jiaming and Hong, Zhibin and Gong, Mingming and Fan, Minhu and Han, Junyu and Liu, Jingtuo and Ding, Errui and Wang, Jingdong},
	year         = 2022,
	month        = {Jun},
	booktitle    = {IEEE/CVF Conference on Computer Vision and Pattern Recognition (CVPR)},
	pages        = {7895--7904},
    doi = {10.1109/CVPR52688.2022.00774}
}

@inproceedings{huang2022agtgan,
	title        = {AGTGAN: Unpaired Image Translation for Photographic Ancient Character Generation},
	author       = {Huang, Hongxiang and Yang, Daihui and Dai, Gang and Han, Zhen and Wang, Yuyi and Lam, Kin-Man and Yang, Fan and Huang, Shuangping and Liu, Yongge and He, Mengchao},
	year         = 2022,
	booktitle    = {ACM International Conference on Multimedia},
	location     = {<conf-loc>, <city>Lisboa</city>, <country>Portugal</country>, </conf-loc>},
	publisher    = {Association for Computing Machinery},
	address      = {New York, NY, USA},
	series       = {MM '22},
	pages        = {5456–5467},
	doi          = {10.1145/3503161.3548338},
	isbn         = 9781450392037                ,
	url          = {https://doi.org/10.1145/3503161.3548338}                ,
	numpages     = 12,
}

@inproceedings{deng2009imagenet,
	title        = {ImageNet: A large-scale hierarchical image database},
	author       = {Deng, Jia and Dong, Wei and Socher, Richard and Li, Li-Jia and Kai Li and Li Fei-Fei},
	year         = 2009,
	booktitle    = {IEEE/CVF Conference on Computer Vision and Pattern Recognition (CVPR)},
	volume       = {},
	number       = {},
	pages        = {248--255},
	doi          = {10.1109/CVPR.2009.5206848},
}

@inproceedings{chen2023textdiffuser,
	title        = {TextDiffuser: Diffusion Models as Text Painters},
	author       = {Chen, Jingye and Huang, Yupan and Lv, Tengchao and Cui, Lei and Chen, Qifeng and Wei, Furu},
	year         = 2023,
	booktitle    = {Advances in Neural Information Processing Systems (NeurIPS)},
	publisher    = {Curran Associates, Inc.},
	volume       = 36,
	pages        = {9353--9387},
	editor       = {A. Oh and T. Neumann and A. Globerson and K. Saenko and M. Hardt and S. Levine}
}

@article{luo2023slogan,
	title        = {SLOGAN: Handwriting Style Synthesis for Arbitrary-Length and Out-of-Vocabulary Text},
	author       = {Luo, Canjie and Zhu, Yuanzhi and Jin, Lianwen and Li, Zhe and Peng, Dezhi},
	year         = 2023,
	journal      = {IEEE Trans. Neural Netw. Learn. Syst.},
	volume       = 34,
	number       = 11,
	pages        = {8503--8515},
	doi          = {10.1109/TNNLS.2022.3151477},
}

@misc{luhman2020diffusion,
	title        = {Diffusion models for Handwriting Generation},
	author       = {Troy Luhman and Eric Luhman},
	year         = 2020,
	eprint       = {2011.06704},
	archiveprefix = {arXiv},
	primaryclass = {cs.LG}
}

@article{gan2022higanplus,
	title        = {{HiGAN+}: Handwriting Imitation {GAN} with Disentangled Representations},
	author       = {Gan, Ji and Wang, Weiqiang and Leng, Jiaxu and Gao, Xinbo},
	year         = 2022,
	month        = {Sep},
	journal      = {ACM Trans. Graph.},
	publisher    = {Association for Computing Machinery},
	address      = {New York, NY, USA},
	volume       = 42,
	number       = 1,
	doi          = {10.1145/3550070 },
	issn         = {0730-0301},
	url          = {https://doi.org/10.1145/3550070}                  ,
	issue_date   = {February 2023},
	articleno    = 11,
	numpages     = 17,
}

@inproceedings{dai2023disentangling,
	title        = {Disentangling Writer and Character Styles for Handwriting Generation},
	author       = {Dai, Gang and Zhang, Yifan and Wang, Qingfeng and Du, Qing and Yu, Zhuliang and Liu, Zhuoman and Huang, Shuangping},
	year         = 2023,
	month        = {Jun},
	booktitle    = {IEEE/CVF Conference on Computer Vision and Pattern Recognition (CVPR)},
	pages        = {5977--5986},
    doi = {10.1109/CVPR52729.2023.00579}
}

@inproceedings{zhu2023diff,
	title        = {Conditional Text Image Generation With Diffusion Models},
	author       = {Zhu, Yuanzhi and Li, Zhaohai and Wang, Tianwei and He, Mengchao and Yao, Cong},
	year         = 2023,
	month        = {Jun},
	booktitle    = {IEEE/CVF Conference on Computer Vision and Pattern Recognition (CVPR)},
	pages        = {14235--14245},
    doi = {10.1109/CVPR52729.2023.01368}
}

@inproceedings{huang2017adain,
	title        = {Arbitrary Style Transfer in Real-Time With Adaptive Instance Normalization},
	author       = {Huang, Xun and Belongie, Serge},
	year         = 2017,
	month        = {Oct},
	booktitle    = {IEEE International Conference on Computer Vision (ICCV)},
    doi = {10.1109/ICCV.2017.167}
}

@article{marti2002iam,
	title        = {The IAM-database: an English sentence database for offline handwriting recognition},
	author       = {Marti, U-V and Bunke, Horst},
	year         = 2002,
	journal      = {Int. J. Doc. Anal. Recog.},
	publisher    = {Springer},
	volume       = 5,
	number       = 1,
	pages        = {39--46},
    doi = {10.1007/s100320200071}
}

@inproceedings{kleber2013cvl,
 author = {Kleber, Florian and Fiel, Stefan and Diem, Markus and Sablatnig, Robert},
 title = {{CVL-DataBase: An Off-Line Database for Writer Retrieval, Writer Identification and Word Spotting}},
 booktitle = {ICDAR},
 year = {2013},
 }

@inproceedings{pippi2023hwd,
	title        = {HWD: A Novel Evaluation Score for Styled Handwritten Text Generation},
	author       = {Vittorio Pippi and Fabio Quattrini and Silvia Cascianelli and Rita Cucchiara},
	year         = 2023,
	booktitle    = {British Machine Vision Conference (BMVC)},
	publisher    = {BMVA},
	url          = {https://papers.bmvc2023.org/0007.pdf}
}

@misc{ding2023improving,
      title={Improving Handwritten OCR with Training Samples Generated by Glyph Conditional Denoising Diffusion Probabilistic Model}, 
      author={Haisong Ding and Bozhi Luan and Dongnan Gui and Kai Chen and Qiang Huo},
      year={2023},
      eprint={2305.19543},
      archivePrefix={arXiv},
      primaryClass={cs.CV},
    url = {https://arxiv.org/abs/2305.19543}
}

@inproceedings{carriere2023detection,
	title        = {Beyond Human Forgeries: An Investigation into Detecting Diffusion-Generated Handwriting},
	author       = {Carri{\`e}re, Guillaume and Nikolaidou, Konstantina and Kordon, Florian and Mayr, Martin and Seuret, Mathias and Christlein, Vincent},
	year         = 2023,
	booktitle    = {International Conference on Document Analysis and Recognition (ICDAR) Workshops},
	publisher    = {Springer Nature Switzerland},
	address      = {Cham},
	pages        = {5--19},
	isbn         = {978-3-031-41498-5 },
	editor       = {Coustaty, Mickael and Forn{\'e}s, Alicia},
    doi = {10.1007/978-3-031-41498-5_1}
}

@inproceedings{lee20202dpe,
	title        = {On Recognizing Texts of Arbitrary Shapes With {2D} Self-Attention},
	author       = {Lee, Junyeop and Park, Sungrae and Baek, Jeonghun and Oh, Seong Joon and Kim, Seonghyeon and Lee, Hwalsuk},
	year         = 2020,
	month        = {Jun},
	booktitle    = {IEEE/CVF Conference on Computer Vision and Pattern Recognition (CVPR) Workshops},
    doi = {10.1109/CVPRW50498.2020.00281}
}

@article{Christlein17PR,
	title        = {Writer Identification Using {GMM} Supervectors and {Exemplar-SVMs}},
	author       = {Christlein, Vincent and Bernecker, David and H{\"{o}}nig, Florian and Maier, Andreas and Angelopoulou, Elli},
	year         = 2017,
	journal      = {Pattern Recognit.},
	volume       = 63,
	pages        = {258--267},
	issn         = {0031-3203},
doi ={10.1016/j.patcog.2016.10.005}
}

@inproceedings{Christlein18DAS,
	title        = {Encoding {CNN} Activations for Writer Recognition},
	author       = {Christlein, Vincent and Maier, Andreas},
	year         = 2018,
	booktitle    = {IAPR International Workshop on Document Analysis Systems},
	address      = {Vienna},
	pages        = {169--174},
    doi={10.1109/DAS.2018.9}
}

@article{Lowe04,
	title        = {Distinctive Image Features from Scale-Invariant Keypoints},
	author       = {Lowe, David G.},
	year         = 2004,
	month        = {Nov},
	journal      = {Int. J. Comput. Vis.},
	volume       = 60,
	number       = 2,
	pages        = {91--110},
	issn         = {0920-5691},
doi = {10.1023/B:VISI.0000029664.99615.94}
}

@inproceedings{Arandjelovic12,
	title        = {Three Things Everyone Should Know to Improve Object Retrieval},
	author       = {Arandjelovi{\'{c}}, R and Zisserman, Andrew},
	year         = 2012,
	month        = {Jun},
	booktitle    = {IEEE/CVF Conference on Computer Vision and Pattern Recognition (CVPR)},
	address      = {Providence},
	pages        = {2911--2918},
doi = {10.1109/CVPR.2012.6248018}
}

@inproceedings{Christlein15ICDAR,
	title        = {Writer Identification Using {VLAD} Encoded Contour-{Zernike} Moments},
	author       = {Christlein, V and Bernecker, D and Angelopoulou, E},
	year         = 2015,
	month        = {Aug},
	booktitle    = {International Conference on Document Analysis and Recognition (ICDAR)},
	address      = {Nancy},
	pages        = {906--910},
doi = {10.1109/ICDAR.2015.7333893}
}

@InProceedings{kodym2021pero,
author="Kodym, Old{\v{r}}ich
and Hradi{\v{s}}, Michal",
editor="Llad{\'o}s, Josep
and Lopresti, Daniel
and Uchida, Seiichi",
title="Page Layout Analysis System for Unconstrained Historic Documents",
booktitle="Document Analysis and Recognition -- ICDAR 2021",
year="2021",
publisher="Springer International Publishing",
address="Cham",
pages="492--506",
doi="10.1007/978-3-030-86331-9_32",
isbn="978-3-030-86331-9"
}

@InProceedings{nauman2024stylus,
author="Riaz, Nauman
and Saifullah, Saifullah
and Agne, Stefan
and Dengel, Andreas
and Ahmed, Sheraz",
editor="Barney Smith, Elisa H.
and Liwicki, Marcus
and Peng, Liangrui",
title="StylusAI: Stylistic Adaptation for Robust German Handwritten Text Generation",
booktitle="Document Analysis and Recognition - ICDAR 2024",
year="2024",
publisher="Springer Nature Switzerland",
address="Cham",
pages="429--444",
isbn="978-3-031-70536-6"
}

\end{document}